%% file: main.tex
\documentclass[letterpaper,journal]{IEEEtran}
\usepackage{amsmath,amsfonts}
\usepackage{array}
\usepackage[caption=false,font=footnotesize]{subfig}
\usepackage{textcomp}
\usepackage{stfloats}
\usepackage{url}
\usepackage{verbatim}
\usepackage{graphicx}
\usepackage{cite}
\input{preamble}

\begin{document}
\title{Parameter-Efficient CLIP Adaptation for 3D Understanding via Unified Tokenization}

\author{
Guofeng Mei\IEEEauthorrefmark{1}\thanks{\IEEEauthorrefmark{1}Guofeng Mei and Qinfeng Xiao contributed equally to this work.},
Qinfeng Xiao\IEEEauthorrefmark{1},
Bin Ren,
Luigi Riz,
Juan Liu,
Xiaoshui Huang,
Xu Zheng, \\
Nicu Sebe, 
Ming-Hsuan Yang,
and Fabio Poiesi
\thanks{G. Mei, L. Riz, and F. Poiesi are with FBK, Trento, Italy. Q. Xiao is with The Hong Kong Polytechnic University, Hong Kong, China. B. Ren is with MBZUAI, Abu Dhabi, UAE. J. Liu is with Beijing Forestry University, Beijing, China. X. Huang is with Shanghai Jiao Tong University, Shanghai, China. X. Zheng is with HKUST, Hong Kong, China. N. Sebe is with the University of Trento, Trento, Italy. M.-H. Yang is with UC Merced, Merced, CA, USA.}
\thanks{Manuscript received July 2026.}
}



\maketitle

\input{secs/0_abstract}
\input{secs/1_intro}
\input{secs/2_related}

\input{secs/3_method}
\input{secs/4_exps}
\input{secs/5_cons}
\bibliographystyle{IEEEtran}
\bibliography{secs/nips}


\input{secs/profile}






\vfill

\end{document}

%% file: preamble.tex
\usepackage{multirow}
\usepackage{amsmath}
\usepackage{algorithmicx}%
\usepackage{algpseudocode}
\usepackage{graphicx}
\usepackage[utf8]{inputenc}      
\usepackage[T1]{fontenc}         
\usepackage[pagebackref=true,breaklinks,colorlinks,citecolor=blue,linkcolor=blue,urlcolor=blue,hypertexnames=false]{hyperref}
\usepackage{booktabs}            
\usepackage{amsfonts}            
\usepackage{nicefrac}            
\usepackage{microtype}           
\usepackage[dvipsnames]{xcolor}  
\usepackage{xspace}
\usepackage{colortbl}
\usepackage{rotating}
\usepackage{array}
\usepackage{wrapfig}
\usepackage[percent]{overpic}
\usepackage{paralist}
\usepackage{enumitem}
\usepackage{todonotes}
\usepackage{arydshln}   

\newcommand{\ourmethod}{UTok3D\xspace}
\newcommand{\knn}{$k$-NN\xspace}
\definecolor{light-gray}{gray}{0.5}
\definecolor{pretty-blue}{RGB}{0, 113, 188}
\definecolor{rowcolor}{gray}{.95} 
\definecolor{linecolor}{gray}{.895} 

\newcommand{\tablefont}{\footnotesize}
\newcommand{\tablearraystretch}{1.05}

\newcommand{\relativeimpP}[1]{\textbf{+#1}}

\newcommand{\relativeimpZ}[1]{0.0}

\def\eg{\emph{e.g.}}
\def\ie{\emph{i.e.}}

%% file: secs/0_abstract.tex
\begin{abstract}
Vision-language models, such as CLIP, encode rich semantic knowledge through large-scale image-text pretraining. 
Reusing these models for 3D understanding is highly desirable, because 3D-text pairs and dense point-level annotations are far scarcer and more difficult to obtain than their 2D counterparts.
However, CLIP is trained on regular 2D image patches, whereas point clouds are unordered, sparse, and irregular. 
The primary challenge lies in constructing 3D token sequences whose geometry, position, and local relations can be interpreted by a frozen, pretrained vision transformer. 
Moreover, coordinate-scale variations across heterogeneous 3D domains make unified training particularly challenging.
To bridge these gaps, we propose a parameter-efficient framework that learns a scale-normalized 3D tokenizer, enabling the CLIP visual encoder to be reused for point-cloud understanding. 
Our tokenizer, named \ourmethod, estimates an input-adaptive geometric scale to calibrate sparse voxelization, coordinate normalization, token-center computation, and 3D positional encoding, and serializes the resulting tokens by Hilbert ordering.
This shared geometric convention makes joint training feasible across point clouds with substantially different metric scales, including object-level shapes, indoor scenes, and outdoor LiDAR scans. The tokenizer is trained without 3D annotations through self-supervised cross-modal distillation from foundation-model features extracted from multi-view images, combining local superpoint alignment with the proposed Sinkhorn Ranked Contrastive distillation. Experiments on ShapeNetPart, ScanNetV2, S3DIS, SemanticKITTI, and nuScenes show that, when equipped with a scale-consistent and CLIP-interpretable token interface, a frozen CLIP visual backbone can be effectively reused for annotation-free 3D segmentation.
\end{abstract}


%% file: secs/1_intro.tex
\section{Introduction}
\label{sec:intro}
\IEEEPARstart{P}{retrained} visual and vision--language models (VLMs), such as CLIP~\cite{radford2021clip}, have emerged as transferable foundation models for a broad range of vision tasks~\cite{caron2021emerging,oquab2023dinov2,tschannen2025siglip}.
Their broad semantic coverage makes them particularly attractive for 3D point-cloud understanding, where large-scale paired 3D--text corpora and dense point-level annotations are substantially more difficult to collect than image--text data~\cite{huang2024frozen,li2026adaptive}. Reusing such pretrained models also provides a promising path toward 3D understanding without training specialized 3D backbones from scratch. However, directly transferring image-trained models to 3D spatial data is non-trivial, as point clouds differ fundamentally from images in both data structure and geometry~\cite{li2026adaptive}. 
One straightforward solution is to fully fine-tune pretrained 2D backbones for each target 3D task or sensing domain, but this incurs substantial parameter, memory, computation, and storage overhead~\cite{li2026adaptive,huang2024frozen}. These limitations motivate parameter-efficient adaptation, where most pretrained weights remain frozen and only compact task- or modality-specific components are learned~\cite{houlsby2019parameter,hu2021lora,jia2022visual,chen2022adaptformer}.

Under this frozen-backbone paradigm, the central question becomes how irregular 3D data should be represented before being processed by the pretrained network. CLIP is trained on image patch tokens, whose spatial layout, local continuity, and semantic statistics are induced by natural images. Point clouds, in contrast, are unordered and irregularly sampled, and they vary significantly in spatial scale, density, and sensing modality~\cite{zhang2026utonia}. 
2D-to-3D transfer requires more than converting point clouds into sequences with the correct input format. It requires constructing 3D tokens whose geometry, relative positions, and local structures can be processed in a manner that better exploits CLIP's pretrained visual knowledge~\cite{li2026adaptive}. 
The 3D tokenizer is not a preprocessing module only, but rather the modality-bridging interface through which frozen image-trained representations can be employed for 3D understanding.

A tokenizer that bridges the 2D--3D modality gap for a single dataset is still insufficient for unified 3D training across heterogeneous domains. Different 3D data sources vary substantially in coordinate scale, spatial extent, point density, and sensing modality~\cite{zhang2026utonia}. Conventional 3D point Transformer tokenization strategies typically construct tokens by sampling coordinate anchors and aggregating local neighborhoods using $k$-nearest neighbors (\knn), ball queries, or voxel grids~\cite{PointBERT,wu2022point,tian2023geomae,zhang2022point,ren2026masked}. While effective in single-dataset supervised or self-supervised settings, these tokenizers often tie token geometry to fixed neighborhood sizes, fixed radii, or dataset-specific coordinate scales. The same tokenization hyperparameter may correspond to substantially different physical extents across domains, such as a small part of an object-level shape~\cite{chang2015shapenet}, a larger region in an indoor RGB-D scene~\cite{dai2017scannet}, or a sparse long-range region in an outdoor LiDAR scan~\cite{behley2019semantickitti}. Without a shared scale-normalized geometric convention, tokens from different domains may encode inconsistent geometric support, making joint training across heterogeneous 3D data less reliable.

These observations suggest that effective frozen-CLIP adaptation to 3D requires a token interface that is both CLIP-interpretable and scale-consistent across domains. Existing methods have explored 2D-to-3D knowledge transfer through multi-view projection, cross-modal feature distillation, and image-pretrained Transformer adaptation~\cite{zhang2022pointclip,zhu2023pointclip,peng2023openscene,umam2024partdistill,qian2024pix4point,huang2024frozen}. Recent unified 3D models further investigate joint training across heterogeneous point-cloud domains using scalable Transformer backbones for 3D scene understanding~\cite{zhang2026utonia}. These methods highlight the value of cross-modal priors and unified 3D training, but they typically train or adapt a 3D encoder, rely on predefined tokenization strategies, or optimize a task-specific 3D backbone. In contrast, we exploit the internal representations of a frozen CLIP visual transformer by learning only a compact 3D tokenizer. The key challenge is to make this interface lightweight while maintaining geometric consistency across domains with different physical scales.

To this end, we introduce \ourmethod, a parameter-efficient framework for adapting frozen CLIP to 3D point clouds through scale-normalized tokenization. Given an input point cloud, \ourmethod estimates an input-adaptive geometric scale from the native point distribution and uses it as a shared unit to calibrate sparse voxelization, coordinate normalization, token-center computation, and 3D positional encoding. The resulting sparse 3D tokens are serialized using Hilbert ordering~\cite{sagan2012space}, forming a deterministic locality-preserving sequence for the frozen visual transformer. This design yields token embeddings with comparable spatial support across domains with different coordinate scales and spatial extents, while concentrating the trainable capacity in a lightweight 3D tokenizer and positional interface with only 3.39M trainable parameters.

The tokenizer is trained without 3D annotations by distilling foundation-model priors from rendered or projected multi-view images. The extracted image features are mapped back to the 3D point cloud and pooled within superpoints, providing efficient and noise-robust teacher representations. A local distillation objective aligns these teachers with the student features produced by the 3D tokenizer and frozen CLIP backbone. To capture broader semantic structure and reduce the effect of noisy or over-smoothed local teacher signals, we further introduce a \emph{Sinkhorn Ranked Contrastive (SRC)} objective. SRC formulates teacher-to-prototype assignment as an entropy-regularized optimal transport~\cite{peyre2019computational} problem with a long-tail-aware prior, producing ranked prototype targets for structural supervision. During inference, token features are aggregated into superpoint-level representations, matched with CLIP text embeddings, and propagated back to points. This enables annotation-free part and semantic segmentation without dataset-specific prediction heads or downstream fine-tuning.

We evaluate \ourmethod across object-level shapes, indoor scenes, and outdoor LiDAR scans on ShapeNetPart~\cite{chang2015shapenet}, ScanNetV2~\cite{dai2017scannet}, S3DIS~\cite{armeni_cvpr16}, SemanticKITTI~\cite{behley2019semantickitti}, and nuScenes~\cite{caesar2020nuscenes}. 
In the main annotation-free setting, a single \ourmethod tokenizer is jointly trained on unlabeled heterogeneous point clouds and directly evaluated on part segmentation for object-level shapes and semantic segmentation for indoor and outdoor scenes without task-specific modification.
Ablation studies show that the proposed input-adaptive scale improves both \knn and voxel tokenizers. Cross-dataset zero-shot transfer experiments further show improved out-of-domain generalization over unnormalized voxel baselines.


%% file: secs/2_related.tex
\section{Related Work}
\label{sec:related-work}
\subsection{2D-to-3D cross-modal knowledge transfer}

Pretrained 2D visual and vision--language models provide rich semantic priors for 3D understanding~\cite{radford2021clip,oquab2023dinov2}. Projection-based methods render point clouds into multi-view images or aggregate image-plane predictions, enabling CLIP-style recognition without learning a native 3D encoder~\cite{zhang2022pointclip,zhu2023pointclip,mei2024geometrically}. However, they often require multi-view rendering or projection at inference time, which can be computationally expensive and may lose fine-grained 3D geometry. 
Subsequent approaches transfer 2D knowledge into 3D through cross-modal feature fusion, knowledge distillation, or open-vocabulary supervision~\cite{peng2023openscene,zhang2023clip,umam2024partdistill,duan2025high,corsetti2026highres,jiang2026proxy3d}. 
Recent open-vocabulary systems, including PGOV3D~\cite{zhang2025pgov3d}, Mosaic3D~\cite{lee2025mosaic3d}, GeoGuide~\cite{tao2026geoguide}, SAL~\cite{ovsep2024better}, OV3D~\cite{jiang2024open}, and LOSC~\cite{samet2026losc}, further improve 3D and LiDAR segmentation, but usually adapt or train specialized 3D segmentation backbones. 

These methods show the value of transferring 2D foundation-model knowledge to 3D, but typically rely on inference-time projection, trainable 3D encoders, or task-specific adaptation. \ourmethod instead uses multi-view foundation features only for annotation-free training and reuses a frozen CLIP visual backbone at inference through a learned 3D token interface.

\subsection{3D tokenization for Vision Transformers}
A common strategy for adapting Vision Transformers (ViTs) to 3D point clouds is to construct local tokens using \knn{}, ball queries, or voxel/grid partitioning~\cite{PointBERT,wu2022point,zhang2022point,tian2023geomae,qi2017pointnet++,yilmaz2026volume,tang2026scenes}. 
These tokenizers preserve local geometry by grouping nearby points or occupied spatial regions. However, their spatial support is often determined by predefined neighborhood sizes~\cite{zhang2022point}, radii~\cite{qi2017pointnet++}, voxel resolutions~\cite{yilmaz2026volume}, or dataset-specific coordinate scales~\cite{zhang2026utonia}. 
As a result, the same tokenization configuration may correspond to substantially different physical extents across heterogeneous domains, making unified training challenging.
Recent adapter-based methods reuse pretrained vision encoders for 3D understanding by replacing image patch embeddings with point-based tokenizers, visual prompts, or modality-specific input modules~\cite{wang2022p2p,qian2024pix4point,huang2024frozen,zhang2023learning}. These methods show that point clouds can be processed by 2D foundation backbones once converted into compatible token sequences, but often require task-specific adaptation, auxiliary 3D backbones, or tokenization schemes sensitive to coordinate-scale changes. In parallel, serialization-based methods explore how discretized 3D structures can be mapped into ordered token sequences for scalable Transformer processing~\cite{wu2024point,wu2025sonata}. However, they are typically designed for trainable 3D networks, where geometric representations and backbone parameters are optimized jointly.

\ourmethod targets a frozen CLIP visual backbone, where the tokenizer must not only construct ordered 3D token sequences but also organize geometry, positions, and local relations in a form compatible with pretrained attention layers. Different from volumetric frameworks such as Volt~\cite{yilmaz2026volume}, which optimize a dedicated 3D Transformer, \ourmethod keeps the CLIP visual backbone frozen and concentrates all trainable parameters in a compact, scale-normalized token interface.

\subsection{Self-supervised 3D representation learning}
\label{subsec:related_ssl}
Self-supervised learning has emerged as a fundamental paradigm for 3D point-cloud representation~\cite{xiao2023unsupervised}. Existing approaches primarily investigate contrastive learning across augmented views~\cite{xie2020pointcontrast,huang2021spatio,rao2020global,liu2022pointdisc,zhang2025self}, clustering-based representation via pseudo-labels or cross-view consistency~\cite{ren2026masked,long2023pointclustering}, and masked modeling for geometric or feature reconstruction~\cite{zhang2022point,yang2023gdmae,chen2023pimae,tian2023geomae}. Additional strategies incorporate autoregressive prediction, serialized tokens, diffusion processes, or state-space models~\cite{cheng2022autoregressive,chen2024pointgpt,yu2026point2seq,liang2024pointmamba,xiao2025pointmadi}. 
Self-supervised dense segmentation frameworks, including LogoSP~\cite{zhang2025logosp}, GrowSP++~\cite{zhang2026growspplusplus}, P-SLCR~\cite{zhan2026pslcr}, and PointGS~\cite{song2026pointgs}, have explored structure-aware grouping and prototype optimization. However, these techniques generally evaluate semantic prediction via post-hoc cluster-to-label alignment rather than enabling direct, head-free matching with textual category embeddings.
Recent works such as Utonia~\cite{zhang2026utonia} train unified 3D encoders across diverse point-cloud domains, while cross-modal pretraining methods transfer 2D model knowledge into trainable 3D encoders~\cite{zhang2026contrastive}.

\ourmethod diverges from Utonia and related pretraining pipelines by avoiding trainable 3D backbones. Instead, it learns only a lightweight, scale-normalized tokenizer that converts point clouds of varying metric scales into CLIP-interpretable token sequences, enabling unified multi-domain training without downstream backbone updates

%% file: secs/3_method.tex
\section{Parameter-Efficient CLIP Adaptation}
\label{sec:method}
\begin{figure*}[t]
    \centering
    \includegraphics[width=\textwidth]{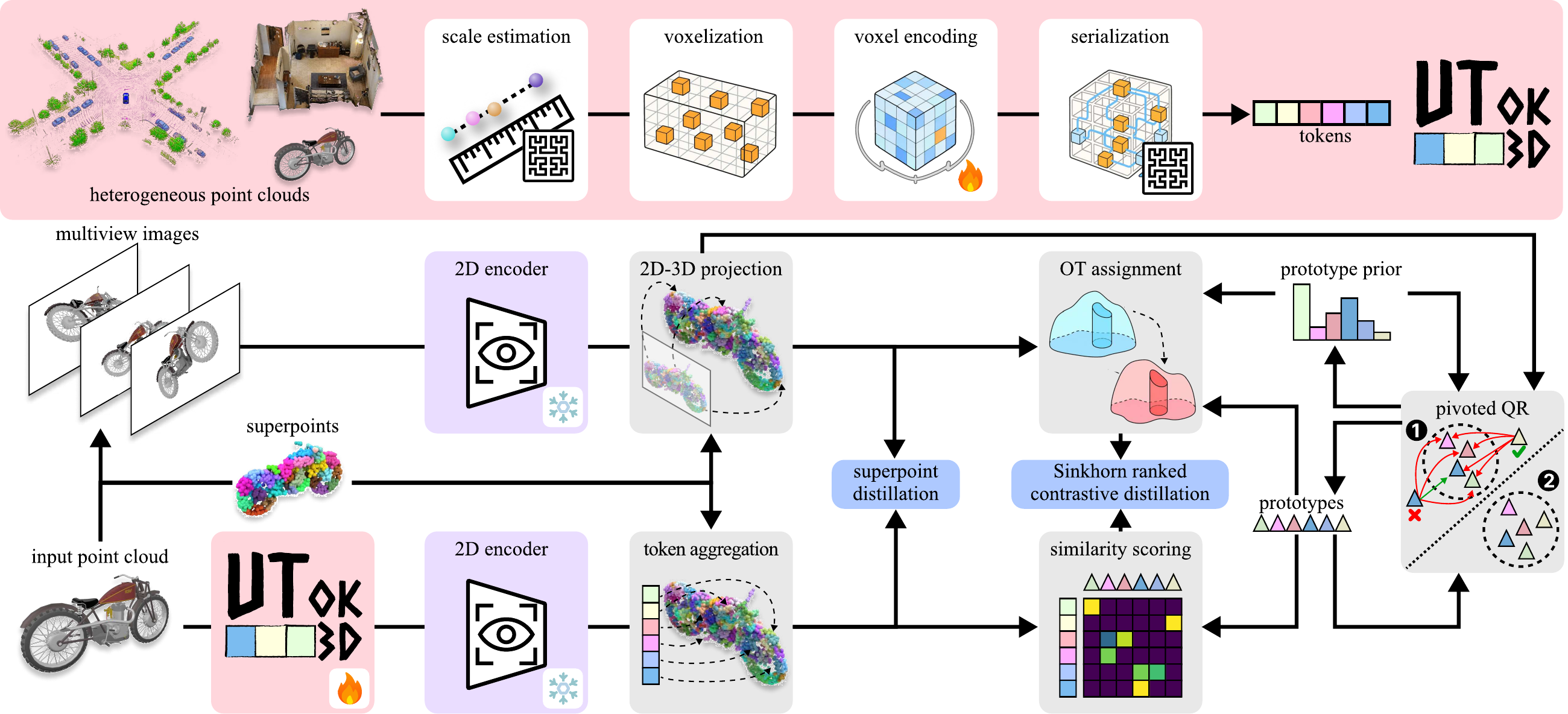}
    \vspace{-6mm}
\caption{Overview of the proposed \ourmethod framework. Given an input point cloud, \ourmethod estimates an input-adaptive geometric scale, constructs scale-normalized sparse voxel tokens, and serializes them via Hilbert ordering for a frozen CLIP visual backbone. The tokenizer is trained without 3D semantic annotations by distilling multi-view features. 
A superpoint-level objective transfers fine-grained 2D--3D correspondence, while Sinkhorn Ranked Contrastive distillation captures broader semantic structure through long-tail-aware prototype assignment. For dense prediction, token features are aggregated into superpoint-level representations and propagated back to points. Symbols: \includegraphics[height=0.9em]{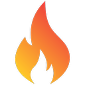} trainable parameters; \includegraphics[height=0.9em]{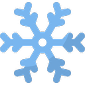} frozen parameters.}
    \label{fig:block_diagram}
    \vspace{-5mm}
\end{figure*}

\ourmethod is a parameter-efficient framework that adapts a frozen CLIP~\cite{radford2021clip} to point clouds through a lightweight,, scale-normalized 3D tokenizer (Fig.~\ref{fig:block_diagram}). 
Given an input point cloud, the tokenizer estimates an input-adaptive geometric scale and uses it to consistently calibrate sparse voxelization, coordinate normalization, token centers, and positional signals. The resulting sparse 3D tokens are serialized by Hilbert ordering~\cite{sagan2012space}, forming a locality-preserving sequence that is fed into the frozen CLIP visual encoder. The encoder outputs are then aggregated into superpoint-level representations, providing a compact interface for both annotation-free training and dense prediction while avoiding a heavy prediction head.

The tokenizer is trained through cross-modal distillation from multi-view foundation-model features. Specifically, a local superpoint-level objective aligns student and teacher features, while Sinkhorn Ranked Contrastive (SRC) distillation transfers broader semantic structure through long-tail-aware ranked prototype assignments. During inference, superpoint-level features are matched with CLIP text embeddings and propagated back to points for annotation-free 3D segmentation.

\subsection{Scale-normalized 3D tokenization}
\label{subsec:tokenization}

Unified 3D tokenization requires that tokens from different domains should retain comparable geometric meaning despite large variations in metric scale and point density. \ourmethod estimates an input-adaptive geometric scale and uses it to calibrate token assignment, coordinate normalization, token-center computation, and positional encoding. This scale-consistent design provides a stable interface between heterogeneous point clouds and frozen visual foundation models such as CLIP.

Let $\mathcal{P}=\{(\mathbf{p}_i,\mathbf{x}_i)\}_{i=1}^{H}$ denote a 3D point cloud, where $\mathbf{p}_i \in \mathbb{R}^{3}$ is the coordinate of point $i$, and $\mathbf{x}_i \in \mathbb{R}^{D_{\mathrm{in}}}$ contains optional attributes such as color, intensity, or normals. We design a tokenizer $\mathcal{T}_{\mathrm{3D}}$ that maps $\mathcal{P}$ into a structured token sequence compatible with a pretrained visual foundation encoder:
\begin{equation}
    \mathcal{T}_{\mathrm{3D}}(\mathcal{P})
    =
    \{(\mathbf{t}_m,\mathbf{o}_m)\}_{m=1}^{M},
    \qquad
    \mathbf{t}_m \in \mathbb{R}^{D},\;
    \mathbf{o}_m \in \mathbb{R}^{3},
\end{equation}
where $\mathbf{t}_m$ and $\mathbf{o}_m$ denote the token feature and scale-normalized token center, respectively. The sequence length $M$ is determined by the number of occupied sparse voxels, allowing the tokenizer to adapt to varying point-cloud extents and sampling densities.

\subsubsection{Scale estimation}
We estimate an input-adaptive scale $\delta$ for subsequent voxelization and coordinate normalization. The points are first sorted by the Hilbert-induced permutation $\pi_{\mathrm{H}}$, yielding a locality-preserving sequence $\mathcal{P}_{\mathrm{H}}=\{\mathbf{p}_{\pi_{\mathrm{H}}(r)}\}_{r=1}^{H}$~\cite{sagan2012space}. We split $\mathcal{P}_{\mathrm{H}}$ into $G$ contiguous chunks $\{\mathcal{I}_g\}_{g=1}^{G}$ and select the middle element of each chunk as its representative, $\bar{\mathbf{p}}_g=\mathbf{p}_{\pi_{\mathrm{H}}(r_g)}$, where $r_g$ is the median ordered index in $\mathcal{I}_g$. The scale is then defined as
\begin{equation}
    \delta
    =
    \operatorname*{median}_{1 \le g < G}
    \left\|
    \bar{\mathbf{p}}_{g+1}
    -
    \bar{\mathbf{p}}_{g}
    \right\|_2 .
\end{equation}
The median provides a robust estimate of typical local spacing by reducing the influence of outlier chunks.

\subsubsection{Sparse voxelization}
Given the estimated scale $\delta$, we set the fine voxel edge length to $e=\delta/K$, where $K$ is a voxelization factor. Let $\boldsymbol{\mu}=\frac{1}{H}\sum_{i=1}^{H}\mathbf{p}_i$ be the point-cloud centroid. Each point is assigned to a fine voxel coordinate
\begin{equation}
    \mathbf{v}^{(0)}_i
    =
    \left\lfloor
    \left({\mathbf{p}_i-\boldsymbol{\mu}}\right)/{e}
    \right\rfloor,
    \qquad
    \mathbf{v}^{(0)}_i \in \mathbb{Z}^{3},
\end{equation}
where $\lfloor\cdot\rfloor$ is applied element-wise. 
The voxel coordinates are then shifted by the minimum coordinate of the input point cloud to obtain non-negative sparse indices.

We use an $L$-level sparse encoder with stride-$2$ downsampling and set $K=2^{L-1}$. With $2^{L-1}e=\delta$, the output voxel coordinate after $L-1$ downsampling stages is $\mathbf{v}^{(L-1)}_i=\left\lfloor\mathbf{v}^{(0)}_i/2^{L-1}\right\rfloor\approx\left\lfloor(\mathbf{p}_i-\boldsymbol{\mu})/\delta\right\rfloor$. Thus, the final token granularity is tied to the adaptive scale $\delta$. The occupied output voxels form the token sequence $\{(\mathbf{t}_m,\mathbf{o}_m)\}_{m=1}^{M}$, where $M=\left|\{\mathbf{v}^{(L-1)}_i\mid \mathbf{p}_i\in\mathcal{P}\}\right|$.
The token number $M$ is given by the number of occupied output voxels. Using the Hilbert-derived scale to set the voxel size, $M$ is typically close to the target granularity $G$, but may vary with the occupancy pattern of each point cloud. Dense or compact regions may merge into fewer occupied voxels, whereas sparse or elongated structures may produce more occupied voxels.


\subsubsection{Voxel encoding}
To reduce cross-domain coordinate discrepancies, we encode voxel features in a scale-normalized frame. Given $\delta$ and a global factor $\alpha$, we normalize each point and compute the normalized token center as
\begin{equation}
    \mathbf{q}_i
    =
    \frac{\mathbf{p}_i-\boldsymbol{\mu}}{\alpha\delta},
    \qquad
    \mathbf{o}_m
    =
    \frac{1}{|V_m|}
    \sum_{i\in V_m}
    \mathbf{q}_i ,
\end{equation}
where $V_m=\{i\mid a_i=m\}$ is the support of token $m$, and $a_i$ denotes the token assignment index for point $i$. For each point $i\in V_m$, we concatenate its normalized position, relative displacement, and optional attributes:
\begin{equation}
    \mathbf{u}_i
    =
    \big[
    \mathbf{q}_i,\;
    \mathbf{q}_i-\mathbf{o}_m,\;
    \mathbf{x}_i
    \big].
\end{equation}
A voxel feature encoder $\Phi$ aggregates these point-wise inputs within each token support:
\begin{equation}
    \mathbf{h}_m
    =
    \Phi\bigl(\{\mathbf{u}_i \mid i\in V_m\}\bigr),
\end{equation}
where $\Phi$ is implemented by point-wise MLPs followed by symmetric pooling. A sparse encoder $\Psi$ then contextualizes the token features using their sparse coordinates:
\begin{equation}
    \{\mathbf{t}_m\}_{m=1}^{M}
    =
    \Psi\bigl(
    \{(\mathbf{h}_m,\mathbf{c}_m)\}_{m=1}^{M}
    \bigr),
\end{equation}
where $\mathbf{c}_m$ denotes the sparse coordinate of token $m$. The output $\mathbf{t}_m$ is the contextualized 3D token feature associated with $\mathbf{o}_m$.

\subsubsection{Hilbert serialization for CLIP}

The CLIP visual transformer requires an ordered token sequence. We therefore sort 3D tokens by the Hilbert rank of their normalized centers:
\begin{equation}
    \pi =
    \operatorname{argsort}
    \bigl(
    \mathcal{H}(\mathbf{o}_1),\dots,\mathcal{H}(\mathbf{o}_M)
    \bigr),
\end{equation}
where $\mathcal{H}(\cdot)$ denotes the Hilbert index.
For the $m$-th ordered token, we add center positional embeddings:
\begin{equation}
    \mathbf{y}_m
    =
    \mathbf{t}_{\pi(m)}
    +
    \phi_{\mathrm{ctr}}
    \bigl(
    \mathbf{o}_{\pi(m)}
    \bigr),
\end{equation}
where $\phi_{\mathrm{ctr}}$ is a lightweight center-encoding MLP. The serialized tokens are processed by the frozen CLIP visual transformer:
\begin{equation}
    [\tilde{\mathbf{Z}}_{\mathrm{cls}},
    \tilde{\mathbf{Z}}_1,\dots,\tilde{\mathbf{Z}}_M]
    =
    \operatorname{Enc}_{\mathrm{CLIP}}
    \bigl(
    [\mathbf{y}_{\mathrm{cls}},
    \mathbf{y}_1,\dots,\mathbf{y}_M]
    \bigr).
\end{equation}
The ordered outputs are mapped back to the original token indices by setting
$\mathbf{Z}_{\pi(m)}=\tilde{\mathbf{Z}}_m$, so that $\mathbf{Z}_m$ denotes the CLIP-enhanced feature of token $m$.

\subsection{Superpoint-level token aggregation}
\label{subsec:superpoint_aggregation}
To avoid memory-intensive storage and noisy point-wise supervision, we aggregate sparse token features into superpoint-level representations via parameter-free pooling.

Let $\ell_i\in\{1,\dots,S\}$ denote the superpoint assignment of point $i$, and let $a_i\in\{1,\dots,M\}$ denote its associated token index. The representation of superpoint $s$ is computed as
\begin{equation}
    \mathbf{g}_s
    =
    \frac{1}{|P_s|}
    \sum_{i\in P_s}
    \mathbf{Z}_{a_i},
    \qquad
    P_s=\{i\mid \ell_i=s\}.
\end{equation}
The superpoint feature is projected into the text-aligned embedding space and normalized as
$\mathbf{e}_s={\psi(\mathbf{g}_s)}/{\|\psi(\mathbf{g}_s)\|_2}$,
where $\psi$ is a lightweight projection head. Given normalized text embeddings
$\{\hat{\mathbf{w}}_c\}_{c=1}^{C}$ for the target categories, the open-vocabulary logits are computed as
$m_{s,c}=\left\langle\mathbf{e}_s,\hat{\mathbf{w}}_c\right\rangle$. The resulting superpoint predictions are broadcast to their constituent points, enabling dense segmentation without a decoder.

\subsection{Annotation-free cross-modal distillation}
\label{subsec:cross_modal_distillation}

We train the tokenizer by distilling multi-view foundation-model features into the 3D token representations. Given a point cloud, the 3D tokenizer and frozen CLIP visual transformer produce superpoint-level student features $\{\mathbf{g}^{\mathrm{st}}_s\}_{s=1}^{S}$. In parallel, multi-view images are processed by a frozen 2D foundation model, and the resulting image features are projected back to the point cloud and fused into teacher superpoint features $\{\mathbf{g}^{\mathrm{te}}_s\}_{s=1}^{S}$ following established protocols~\cite{jiang2024open,zhu2022pointclip2}. Both student and teacher features are $L_2$-normalized. The training loss is
\begin{equation}
    \mathcal{L}_{\mathrm{dist}}
    =
    \lambda_{\mathrm{sd}}\mathcal{L}_{\mathrm{sd}}
    +
    \lambda_{\mathrm{SRC}}\mathcal{L}_{\mathrm{SRC}},
\end{equation}
where $\mathcal{L}_{\mathrm{sd}}$ performs paired superpoint alignment and $\mathcal{L}_{\mathrm{SRC}}$ transfers semantic structure via ranked prototype supervision.

\subsubsection{Superpoint distillation}

We align student and teacher superpoint features via cosine distance:
\begin{equation}
    \mathcal{L}_{\mathrm{sd}}
    =
    \frac{1}{S}
    \sum_{s=1}^{S}
    \left(
    1-
    \left\langle
    \mathbf{g}^{\mathrm{st}}_s,
    \mathbf{g}^{\mathrm{te}}_s
    \right\rangle
    \right).
\end{equation}

\subsubsection{Sinkhorn ranked contrastive distillation}

Superpoint distillation aligns paired 2D--3D features but overlooks semantic relations across regions, especially under noisy multi-view projection and long-tailed 3D distributions. We introduce Sinkhorn Ranked Contrastive (SRC) distillation, which transfers global teacher structure through long-tail-aware prototype assignment and high-confidence ranked supervision.

\paragraph{Codebook construction}
We maintain a buffer of normalized teacher features $\mathcal{B}=\{\mathbf{z}_n\}_{n=1}^{N_b}$, where $\|\mathbf{z}_n\|_2=1$. 
We construct a codebook $\mathcal{C}=\{\boldsymbol{\kappa}_j\}_{j=1}^{J}$ via residual-pivoted QR sampling. 
Starting from $\mathbf{r}^{(0)}_n=\mathbf{z}_n$, we select
$n_j=\arg\max_n\|\mathbf{r}^{(j-1)}_n\|_2$ and
$\mathbf{q}_j=\mathbf{r}^{(j-1)}_{n_j}/\|\mathbf{r}^{(j-1)}_{n_j}\|_2$, and update
\begin{equation}
    \mathbf{r}^{(j)}_n
    =
    \mathbf{r}^{(j-1)}_n
    -
    \langle
    \mathbf{r}^{(j-1)}_n,\mathbf{q}_j
    \rangle
    \mathbf{q}_j .
\end{equation}
The selected feature is used as the prototype, \ie, $\boldsymbol{\kappa}_j=\mathbf{z}_{n_j}$, and the codebook is fixed during student training.

\paragraph{Long-tail prototype prior}
Each buffered feature is assigned to its nearest prototype,
$b_n=\arg\max_j\langle \mathbf{z}_n,\boldsymbol{\kappa}_j\rangle$.
The prototype frequency and its mean are computed as
\begin{equation}
    f_j=
    \frac{1}{N_b}
    \sum_{n=1}^{N_b}
    \mathbb{I}(b_n=j),
    \qquad
    \bar{f}=
    \frac{1}{J}
    \sum_{j=1}^{J}f_j .
\end{equation}
To suppress frequent prototypes, we define
$r_j=(\bar{f}+\epsilon)/(f_j+\epsilon)$ and calculate the prototype prior $\boldsymbol{\rho}=\{\rho_j\}_{j=1}^J$ by 
\begin{equation}
\rho_j =\frac{\tilde{w}_j}{\sum_{\ell=1}^{J}\tilde{w}_{\ell}}, \tilde{w}_j=\operatorname{clip}_{[w_l,w_r]}\left(1+\gamma\log\frac{1+r_j}{2}\right).
\end{equation}
Here, $\gamma$ controls the balancing strength, $\epsilon=10^{-4}$ ensures numerical stability, and $[w_l,w_r]$ bounds the prototype weights. 

\paragraph{Sinkhorn target assignment}
Given teacher superpoint features, we compute prototype affinities
$A_{sj}=\langle \mathbf{g}^{\mathrm{te}}_s,\boldsymbol{\kappa}_j\rangle$.
The teacher assignment matrix is obtained by optimal transport~\cite{peyre2019computational}:
\begin{equation}
\begin{aligned}
    \mathbf{Q}^{*}
    &=
    \arg\max_{\mathbf{Q}\in\mathcal{U}(\mathbf{a},\boldsymbol{\rho})}
    \langle \mathbf{Q},\mathbf{A}\rangle
    +
    \tau_t\mathcal{H}(\mathbf{Q}), \\
    \mathcal{U}(\mathbf{a},\boldsymbol{\rho})
    &=
    \left\{
    \mathbf{Q}\ge 0
    \mid
    \mathbf{Q}\mathbf{1}_{J}=\mathbf{a},
    \;
    \mathbf{Q}^{\top}\mathbf{1}_{S}=\boldsymbol{\rho}
    \right\},
\end{aligned}
\end{equation}
where $\mathcal{H}(\mathbf{Q})=-\sum_{s,j}Q_{sj}\log Q_{sj}$ and
$\mathbf{a}=S^{-1}\mathbf{1}_{S}$. The non-uniform prior balances prototype usage.

\paragraph{Ranked target truncation and SRC loss}
To remove low-confidence assignments, we retain only the top-$R$ prototypes
$\Omega_s$ for each superpoint and renormalize the transport scores. This gives the ranked target $Q^{\mathrm{rk}}_{sj}$, while the student prediction over teacher prototypes is computed as
\begin{equation}
Q^{\mathrm{rk}}_{sj}=\frac{Q^{*}_{sj}\mathbb{I}(j\in\Omega_s)}
    {\sum_{\ell\in\Omega_s}Q^{*}_{s\ell}},
    P_{sj}
    =
\frac{\exp\left(\langle\mathbf{g}^{\mathrm{st}}_s,\boldsymbol{\kappa}_j\rangle/\tau_s\right)}
{\sum_{\ell=1}^{J}\exp\left(\langle\mathbf{g}^{\mathrm{st}}_s,\boldsymbol{\kappa}_{\ell}\rangle/\tau_s\right)} .
\end{equation}
The SRC loss is the cross-entropy between ranked teacher targets and student predictions:
\begin{equation}
    \mathcal{L}_{\mathrm{SRC}}
    =
    -\frac{1}{S}
    \sum_{s=1}^{S}
    \sum_{j=1}^{J}
    Q^{\mathrm{rk}}_{sj}\log P_{sj}.
\end{equation}
By combining long-tail-aware Sinkhorn assignment with ranked truncation, SRC provides balanced structural supervision while suppressing noisy teacher targets.

%% file: secs/4_exps.tex
\section{Experiments}

\subsection{Experimental setup}
\label{subsec:experimental_setup}

We assess \ourmethod across object-level shapes, indoor scenes, and outdoor LiDAR point clouds. The evaluation spans five standard benchmarks with diverse object scales, spatial layouts, point densities, and sensing modalities:
\begin{itemize}[leftmargin=*, labelsep=0.5em]
    \item \textit{ShapeNetPart~\cite{chang2015shapenet}:} Used for object-level part segmentation, with 2,874 test shapes from 16 object categories and 50 part labels. Following the standard protocol~\cite{mei2024geometrically,zhu2023pointclip}, each instance is uniformly sampled to 2,048 points, and 10 multi-view images are rendered per shape. We report class-averaged mean intersection over union (mIoU, \%).
    \item \textit{ScanNetv2~\cite{dai2017scannet} \& S3DIS~\cite{armeni_cvpr16}:} Employed for indoor semantic segmentation. ScanNetv2 contains 1,513 indoor scenes with 20 semantic classes, and we report results on the validation split. S3DIS contains 3D scans of 271 rooms across six areas, and we evaluate on Area 5 with 13 semantic classes. We report mIoU and mean accuracy (mAcc, \%).
    \item \textit{SemanticKITTI~\cite{behley2019semantickitti} and nuScenes~\cite{caesar2020nuscenes}:} Used to evaluate scalability to large-scale outdoor LiDAR semantic segmentation. We report mIoU on both benchmarks.
\end{itemize}

\ourmethod focuses on annotation-free 3D segmentation by reusing a frozen CLIP visual backbone and learning only a lightweight 3D tokenizer; we select representative baselines for each evaluation task. Fully supervised 3D backbones are reported only as upper-reference results.
\begin{itemize}[leftmargin=*, labelsep=0.5em]
    \item \textit{Part segmentation:} We compare with multi-view VLM projection methods, including PointCLIP~\cite{zhang2022pointclip}, PointCLIPv2~\cite{zhu2023pointclip}, and GeoZe~\cite{mei2024geometrically}, as well as PartDistill~\cite{umam2024partdistill}, which distills VLM knowledge into a trainable 3D backbone.
    
    \item \textit{Indoor semantic segmentation:} We compare with open-vocabulary projection and distillation methods, including OpenScene~\cite{peng2023openscene}, CLIP-FO3D~\cite{zhang2023clip}, PGOV3D~\cite{zhang2025pgov3d}, Mosaic3D~\cite{lee2025mosaic3d}, and GeoGuide~\cite{tao2026geoguide}. We also include unsupervised clustering-based methods that learn task-specific 3D representations, such as LogoSP~\cite{zhang2025logosp}, GrowSP++~\cite{zhang2026growspplusplus}, P-SLCR~\cite{zhan2026pslcr}, and PointGS~\cite{song2026pointgs}, but report them separately because they require dataset-specific cluster-to-label alignment during evaluation.
    
    \item \textit{Outdoor LiDAR semantic segmentation:} We compare with recent open-vocabulary LiDAR segmentation methods, including OpenScene~\cite{peng2023openscene}, SAL~\cite{ovsep2024better}, OV3D~\cite{jiang2024open}, and LOSC~\cite{samet2026losc}.
\end{itemize}

We use CLIP-ViT-B/16~\cite{radford2021clip} as the foundation architecture. The original 2D image patch tokenizer and positional embeddings are replaced by the proposed scale-normalized 3D tokenizer and 3D positional encoding, while both the CLIP visual transformer and text encoder remain frozen. The text encoder provides category-level textual embeddings for open-vocabulary evaluation.
For the primary annotation-free setting, a single 3D tokenizer is jointly trained on heterogeneous unlabeled datasets, including ShapeNetPart, ScanNetv2, S3DIS, nuScenes, and SemanticKITTI, without using 3D semantic annotations. The trained tokenizer is then directly evaluated on part and semantic segmentation tasks without downstream fine-tuning, parameter updates, or dataset-specific adaptation.

\subsection{Implementation details}
\label{subsec:implementation_details}

\ourmethod is implemented in PyTorch~\cite{paszke2019pytorch} and trained on four NVIDIA A100 GPUs, each with 64,GB memory. We jointly train \ourmethod for 100 epochs using AdamW~\cite{loshchilov2017decoupled} with a total batch size of 32, an initial learning rate of $5\times10^{-5}$ under cosine decay, and a weight decay of $0.05$.
The trainable tokenizer contains 3.39M parameters, mainly from MLP modules.

The input-adaptive geometric scale $\delta$ is estimated per sample from $G=384$ anchor centers. We use an $L$-level sparse encoder with stride-$2$ downsampling and voxelization factor $K=2^{L-1}$, with $L=2$, $K=2$, and $\alpha=1$ by default. The token length $M$ varies with sparse voxel occupancy but remains close to $G$ on average. Token sequences are batch-padded and processed with a validity mask.
NACLIP~\cite{hajimiri2025pay} is applied to extract teacher features from multi-view images. 
We use the dataset-provided superpoints on ScanNetv2 and generate superpoints on S3DIS, SemanticKITTI, and nuScenes using $L_0$-cut~\cite{landrieu2018large}.
The training objective combines local superpoint distillation and global SRC distillation, with $\lambda_{\mathrm{sd}}=1.0$ and $\lambda_{\mathrm{SRC}}=0.1$. For SRC, the teacher feature buffer size is $N_b=65536$, and the codebook $J$ determined by the teacher features. The ranked target keeps the top-$R$ assignments, with $R=4$ by default. The teacher Sinkhorn temperature and student contrastive temperature are $\tau_t=0.05$ and $\tau_s=0.1$, respectively. For the long-tail prior, we use $\gamma=0.1$, clipping range $[w_l,w_r]=[0.2,5.0]$, and three Sinkhorn iterations.

\subsection{Annotation-free segmentation}
\label{subsec:exp:annfree_seg}
We evaluate \ourmethod under an annotation-free training paradigm with a single jointly trained tokenizer. Following Utonia~\cite{zhang2026utonia}, we use a unified input interface by concatenating 3D coordinates, color channels, and surface normals, with zero vectors for missing modalities. During joint training, optional color and normal channels are randomly blinded while coordinates are always retained. This masking strategy improves robustness to missing, noisy, or domain-shifted appearance and geometric attributes.
The trained tokenizer is directly evaluated on part and semantic segmentation tasks without downstream fine-tuning. For all segmentation tasks, inference is head-free: we compute cosine similarities between learned superpoint-level 3D features and text embeddings of candidate labels, including part labels for ShapeNetPart and semantic category labels for scene-level benchmarks. The predicted superpoint labels are then propagated back to individual points for evaluation. This protocol tests whether the proposed scale-normalized tokenizer can provide a unified 3D token interface across heterogeneous 3D domains.

\subsubsection{Part segmentation on ShapeNetPart}
\label{subsec:exp:partseg}
\input{table/shapenet}
\input{fig_tex/partseg}
We first evaluate the zero-shot capacity of the jointly trained tokenizer on open-vocabulary part segmentation using the ShapeNetPart benchmark~\cite{chang2015shapenet}. 
Tab.~\ref{tab:part_seg} shows that \ourmethod achieves 59.3\% mIoU under this zero-shot, head-free cosine matching setting. When compared directly against our reproduced PointCLIPv2$^\dagger$~\cite{zhu2023pointclip} baseline, \ourmethod yields an absolute improvement of +7.5 mIoU points (from 51.8\% to 59.3\%). It additionally outperforms GeoZe~\cite{mei2024geometrically} by +1.9 mIoU points, indicating that an optimized 3D tokenizer captures more precise, fine-grained part structures than direct multi-view feature projection. 
Compared with PartDistill~\cite{umam2024partdistill}, which distills VLM knowledge into a trainable 3D backbone, \ourmethod improves mIoU by +5.5 points while keeping the CLIP-ViT-B/16 visual backbone frozen and requiring no task-specific fine-tuning.
Per-category analysis reveals that \ourmethod achieves the highest mIoU across 9 out of 16 categories, including \textit{Airplane}, \textit{Bag}, \textit{Car}, \textit{Chair}, \textit{Guitar}, \textit{Knife}, \textit{Laptop}, \textit{Pistol}, and \textit{Table}. These performance gains are particularly pronounced in categories governed by stable, distinct geometric primitives (\eg, \textit{Bag}, \textit{Chair}, \textit{Laptop}, and \textit{Table}). Conversely, performance exhibits minor limitations in categories such as \textit{Lamp}, \textit{Motorbike}, and \textit{Skateboard}, where fine-grained boundaries are thin, sparse, or inherently ambiguous under direct open-vocabulary text matching. This underscores the capacity of the proposed tokenizer to ground textual semantics into 3D shapes. Qualitative segmentation comparisons are detailed in Fig.~\ref{fig:partseg}, highlighting more coherent part region boundaries under our \ourmethod, alongside representative failure modes on highly intricate structures.

\subsubsection{Semantic segmentation on indoor scenes}
\label{subsec:exp:seman_seg}
\input{table/indoor}
\input{fig_tex/indoor}

We assess zero-shot indoor semantic segmentation across the ScanNetv2~\cite{dai2017scannet} and S3DIS~\cite{armeni_cvpr16} datasets. 
Tab.~\ref{tab:clip_sem_seg} compares \ourmethod with multi-view VLM projection baselines, unsupervised clustering methods, and 3D cross-modal distillation frameworks. The ``Backbone / Rep.'' column distinguishes methods by their underlying representations, including multi-view projection, sparse convolutional networks, 3D Gaussian splatting, and frozen visual transformers. 
\texttt{pcd} and \texttt{img} denote the input point cloud and multi-view images, respectively.
Unsupervised clustering methods require dataset-dependent Hungarian label alignment; they are listed separately from zero-shot open-vocabulary text-matching methods. In contrast, \ourmethod directly matches learned 3D features with category text embeddings and applies the same jointly trained tokenizer to both benchmarks without downstream task adaptation.
On the ScanNetv2 validation set, \ourmethod achieves 59.2\% mIoU and 72.4\% mAcc. This performance consistently surpasses multi-view projection frameworks like OpenScene~\cite{peng2023openscene} and GeoZe~\cite{mei2024geometrically}, indicating that a dedicated 3D token interface generates scene-level representations superior to ungrounded 2D projected features. It similarly outperforms CLIP-FO3D~\cite{zhang2023clip} and OpenScene's sparse convolutional branch in mIoU, despite using a frozen CLIP-ViT-B/16 backbone with a lightweight 3.39M-parameter tokenizer. While PGOV3D~\cite{zhang2025pgov3d} and GeoGuide~\cite{tao2026geoguide} yield slightly higher mIoU scores, they necessitate fully trainable 3D segmentation backbones or extensive 3D pretraining phases. Conversely, \ourmethod confines all optimization to the tokenizer, leaving the CLIP visual transformer unmodified. On S3DIS Area 5, \ourmethod achieves 49.7\% mIoU and 61.1\% mAcc, outperforming existing zero-shot text-matching baselines in Tab.~\ref{tab:clip_sem_seg}. These findings confirm that a frozen 2D visual foundation model can be successfully repurposed for dense 3D semantic prediction when equipped with a scale-consistent token interface.
Fig.~\ref{fig:scannet_s3dis} compares open-vocabulary semantic segmentation on ScanNetv2 and S3DIS. The left three columns show ScanNetv2 results, while the right three columns show S3DIS results. Compared with OpenScene and GeoZe, \ourmethod produces more accurate and coherent predictions.

\subsubsection{Semantic segmentation on outdoor LiDAR scenes}
\label{subsec:exp:outdoor_lidar}
We further assess annotation-free semantic segmentation on SemanticKITTI~\cite{behley2019semantickitti} and nuScenes~\cite{caesar2020nuscenes}. Compared with object-level shapes and indoor RGB-D scenes, outdoor LiDAR scans exhibit different sensing patterns, including long-range sparsity, non-uniform point density, and large scene-scale variation. 
Tab.~\ref{tab:semantickitti_main} reports the results. \ourmethod achieves 49.1\% mIoU on nuScenes and 34.3\% mIoU on SemanticKITTI, showing effective transfer to outdoor LiDAR scenes. It performs competitively with recent 3D-backbone distillation methods while keeping the CLIP visual transformer frozen and training only a lightweight 3D tokenizer. These results indicate that input-adaptive scale normalization helps reduce sensitivity to dataset-specific metric scales and provides a stable token interface across heterogeneous 3D domains.
Fig.~\ref{fig:kitti} provides open-vocabulary semantic segmentation on SemanticKITTI.
\begin{table}[t]
    \centering
    \tablefont
    \renewcommand{\arraystretch}{\tablearraystretch}
    \setlength{\tabcolsep}{2pt}
    \caption{Annotation-free semantic segmentation on outdoor LiDAR benchmarks. We report mIoU (\%) for nuScenes and SemanticKITTI. ``--'' indicates unavailable results.}
    \label{tab:semantickitti_main}
    \begin{tabular}{lcccc}
    \toprule
    Method & Venue & Supervision & nuScenes & SemanticKITTI \\
    \midrule
    \multicolumn{5}{c}{\textit{3D-backbone distillation-based methods}} \\
    \midrule
    OpenScene~\cite{peng2023openscene} & CVPR 2023 & LSeg~\cite{li2022languagedriven} & 42.1 & --\\
    SAL~\cite{ovsep2024better}          & ECCV 2024 & CLIP~\cite{radford2021clip} & 33.9 & 28.7 \\
    OV3D~\cite{jiang2024open}           & CVPR 2024 & CLIP~\cite{radford2021clip} & 45.5 & -- \\
    LOSC~\cite{samet2026losc}           & 3DV 2026  & OpenSeeD~\cite{zhang2023simple} & \bf49.3 & \bf35.2 \\
    \midrule
    \multicolumn{5}{c}{\textit{Tokenizer distillation-based method}} \\
    \midrule
    \rowcolor{rowcolor}\ourmethod & Ours & CLIP~\cite{radford2021clip} & 49.1 & 34.3 \\
    \bottomrule
    \end{tabular}
\end{table}
\input{fig_tex/kitti}

\subsection{Ablation study and analysis}
\label{subsec:ablations}

We conduct ablation studies and analyses to examine the main design components of \ourmethod. Unless otherwise stated, these experiments use independent single-dataset training.

\subsubsection{Tokenizer topology and scale normalization}
\label{subsubsec:tokenizer_topology}

We first analyze the effect of tokenizer topology and scale normalization (SN). We compare a standard \knn tokenizer~\cite{PointBERT} with our sparse voxel tokenizer on three geometrically different domains: ShapeNetPart for object-level shapes, ScanNetv2 for indoor scenes, and SemanticKITTI for outdoor LiDAR scans. For each tokenizer, we evaluate variants without and with scale normalization, denoted by \textit{w/o SN} and \textit{w/ SN}, respectively. SN maps raw 3D coordinates into an input-adaptive normalized coordinate frame before token encoding.

Tab.~\ref{tab:sec:norm} shows that scale normalization improves both tokenizer types across all three benchmarks, with the largest gains on SemanticKITTI. For the \knn tokenizer, SN improves mIoU from 57.98\% to 58.22\% on ShapeNetPart, from 55.94\% to 56.21\% on ScanNetv2, and from 24.49\% to 28.63\% on SemanticKITTI. The corresponding mAcc also increases from 69.35\% to 69.49\%, from 66.23\% to 67.09\%, and from 36.93\% to 43.66\%, respectively. This indicates that input-adaptive normalization is especially important for outdoor LiDAR scans, where coordinate ranges and sampling densities vary substantially.
The sparse voxel tokenizer also benefits from SN. Its mIoU improves from 58.05\% to 58.97\% on ShapeNetPart, from 55.74\% to 56.79\% on ScanNetv2, and from 30.79\% to 34.26\% on SemanticKITTI. Under the same SN setting, the voxel tokenizer outperforms the \knn tokenizer by +0.75, +0.58, and +5.63 mIoU points on ShapeNetPart, ScanNetv2, and SemanticKITTI, respectively. These results indicate that scale-normalized sparse voxelization provides more stable spatial support than fixed-count neighborhood grouping, especially for large-scale outdoor scenes. We therefore use the scale-normalized voxel tokenizer as the default design.

\begin{table}[t]
    \centering
    \tablefont
    \renewcommand{\arraystretch}{\tablearraystretch}
    \setlength{\tabcolsep}{5pt}
    \caption{Quantitative comparison of tokenization topologies under the same training protocol on ShapeNetPart, ScanNetv2, and SemanticKITTI. Bold denotes the best result in each column. SN indicates scale normalization.}
    \label{tab:sec:norm}
    \begin{tabular}{lcccccc}
    \toprule
    \multirow{2}{*}{Method}  & \multicolumn{2}{c}{ShapeNetPart} & \multicolumn{2}{c}{ScanNetv2} & \multicolumn{2}{c}{SemanticKITTI} \\
    \cmidrule(lr){2-3} \cmidrule(lr){4-5} \cmidrule(lr){6-7}
                             & mIoU$\uparrow$ & mAcc$\uparrow$ & mIoU$\uparrow$ & mAcc$\uparrow$ & mIoU$\uparrow$ & mAcc$\uparrow$ \\
    \midrule
    \multicolumn{7}{c}{\textit{$k$-NN-based tokenizer}} \\
    \midrule
    w/o SN                   & 57.98 & 69.35 & 55.94 & 66.23 & 24.49 & 36.93  \\
    w/ SN                    & 58.22 & 69.49 & 56.21 & 67.09 & 28.63 & 43.66  \\
    \midrule
    \multicolumn{7}{c}{\textit{Voxel-based tokenizer}} \\
    \midrule
    w/o SN                   & 58.05 & 69.66 & 55.74 & 67.02 & 30.79 & 42.04  \\
    \rowcolor{rowcolor}w/ SN & \textbf{58.97} & \textbf{69.93} & \textbf{56.79} & \textbf{67.81} & \textbf{34.26} & \textbf{48.92} \\
    \bottomrule
    \end{tabular}
\end{table}

\subsubsection{Cross-dataset generalization}
\label{subsubsec:cross_dataset_generalization}

We further explore whether the scale-normalized tokenizer generalizes across datasets without downstream fine-tuning. The tokenizer is pretrained with annotation-free distillation on a source domain and directly transferred to an unseen target domain. We consider two cross-domain settings: ScanNetv2 $\rightarrow$ S3DIS for indoor scene transfer and SemanticKITTI $\rightarrow$ nuScenes for outdoor LiDAR transfer.

Tab.~\ref{tab:cross_generalization} shows that the voxel tokenizer without scale normalization degrades substantially under domain shift, achieving only 22.7\% mIoU on ScanNetv2 $\rightarrow$ S3DIS and 7.3\% mIoU on SemanticKITTI $\rightarrow$ nuScenes. This indicates that fixed unnormalized voxel grids are sensitive to changes in scene scale, point density, and sensor characteristics. With scale normalization, \ourmethod improves transfer performance to 26.9\% mIoU on S3DIS and 10.1\% mIoU on nuScenes, corresponding to absolute gains of +4.2 and +2.8 mIoU points, respectively. These results show that the Hilbert-driven adaptive scale $\delta$ helps standardize token geometry across datasets, providing a more consistent interface to the frozen CLIP backbone.

\begin{table}[t]
    \centering
    \tablefont
    \renewcommand{\arraystretch}{\tablearraystretch}
    \setlength{\tabcolsep}{3pt}
    \caption{Cross-dataset generalization of \ourmethod. We report mIoU (\%), and $\Delta$ mIoU is the absolute gain over the voxel tokenizer without scale normalization.}
    \label{tab:cross_generalization}
    \begin{tabular}{lcccc}
    \toprule
    \multirow{2}{*}{Method} & \multicolumn{2}{c}{ScanNetv2 $\rightarrow$ S3DIS}  & \multicolumn{2}{c}{SemanticKITTI $\rightarrow$ nuScenes}\\
    \cmidrule(lr){2-3} \cmidrule(lr){4-5}
                            & mIoU$\uparrow$ & $\Delta$ mIoU$\uparrow$ & mIoU$\uparrow$ & $\Delta$ mIoU$\uparrow$ \\
    \midrule
    voxel w/o SN            & 22.7 & --           & 7.3  & -- \\
    \rowcolor{rowcolor}\ourmethod w/ SN & \textbf{26.9} & \relativeimpP{4.2} & \textbf{10.1} & \relativeimpP{2.8} \\
    \bottomrule
    \end{tabular}
\end{table}

\subsubsection{Cross-modal distillation objectives}
\label{subsubsec:ablation_objective}

We examine the contribution of each training objective. The full objective combines local superpoint distillation with global Sinkhorn Ranked Contrastive (SRC) distillation. Superpoint distillation preserves paired 2D--3D correspondence by aligning each student superpoint with its corresponding multi-view teacher feature, whereas SRC transfers global semantic structure through ranked prototype supervision.

Tab.~\ref{tab:ablation_objective} shows that the two objectives are complementary. Using only superpoint distillation or only SRC distillation achieves 57.9\% and 57.2\% mIoU on ShapeNetPart, respectively. Replacing SRC with a uniform Sinkhorn target improves performance to 58.6\% mIoU, but remains below the full formulation. Combining superpoint distillation with long-tail-aware SRC achieves the best performance of 59.0\% mIoU, validating the importance of jointly enforcing local feature alignment and global prototype-level structural supervision.

\begin{table}[t]
    \centering
    \tablefont
    \renewcommand{\arraystretch}{\tablearraystretch}
    \setlength{\tabcolsep}{4pt}
    \caption{Ablation of annotation-free training objectives on ShapeNetPart. We report instance-average mIoU (\%), and $\Delta$ mIoU is measured against superpoint distillation only.}
    \label{tab:ablation_objective}
    \begin{tabular}{lcc}
    \toprule
    Training objective & mIoU$\uparrow$ & $\Delta$mIoU$\uparrow$ \\
    \midrule
    Superpoint distillation only & 57.9 & -- \\
    SRC distillation only        & 57.2 & -- \\
    Superpoint distillation + uniform Sinkhorn & 58.6 & +0.7 \\
    \rowcolor{rowcolor}Superpoint distillation + SRC (ours) & \textbf{59.0} & \textbf{+1.1} \\
    \bottomrule
    \end{tabular}
\end{table}

\subsubsection{Codebook construction}
\label{subsubsec:ablation_codebook}

The SRC loss relies on a global codebook of semantic anchors. We compare different codebook construction strategies, including random prototypes, spherical $k$-means, threshold-based leader clustering, and residual-pivoted codebook generation.
Tab.~\ref{tab:ablation_codebook} shows that codebook construction affects cross-modal token learning. Random prototypes achieve 57.7\% mIoU, while spherical $k$-means obtains 57.4\% mIoU. This suggests that centroid-based clustering may overemphasize high-density teacher feature regions under imbalanced or long-tailed distributions, producing redundant anchors that poorly represent minority semantic modes. Threshold-based leader clustering improves performance to 58.2\% mIoU by adaptively discovering feature modes. Our residual-pivoted codebook achieves the best result of 59.0\% mIoU by selecting anchors with large orthogonal residuals, promoting prototype diversity and broader coverage of the teacher embedding space.

\begin{table}[t]
    \centering
    \tablefont
    \renewcommand{\arraystretch}{\tablearraystretch}
    \setlength{\tabcolsep}{8pt}
    \caption{Ablation of codebook construction strategies for SRC distillation. We report mIoU (\%) on ShapeNetPart, and $\Delta$ mIoU is measured against random prototypes.}
    \label{tab:ablation_codebook}
    \begin{tabular}{lcc}
    \toprule
    Codebook construction & mIoU$\uparrow$ & $\Delta$mIoU$\uparrow$ \\
    \midrule
    Random prototypes & 57.7 & -\\
    Spherical $k$-means & 57.4 & -0.3 \\
    Threshold-based leader clustering & 58.2 & +0.5 \\
    \rowcolor{rowcolor}Residual-pivoted codebook & \bf59.0 & \bf+1.3\\
    \bottomrule
    \end{tabular}
\end{table}

\subsubsection{Ranked target sparsity in SRC}
\label{subsubsec:ablation_topr}

We investigate the sparsity of SRC supervision by varying the number of retained top-$R$ ranked targets. Tab.~\ref{tab:ablation_topm} shows that performance increases from 57.9\% mIoU at $R=1$ to 59.0\% mIoU at $R=4$, indicating that retaining multiple high-confidence prototypes better captures local semantic ambiguity than a single target. Increasing $R$ beyond 4 or retaining all prototypes slightly reduces performance, likely because lower-ranked assignments introduce less discriminative supervision. We therefore set $R=4$ as the default choice.

\begin{table}[t]
    \centering
    \tablefont
    \renewcommand{\arraystretch}{\tablearraystretch}
    \setlength{\tabcolsep}{8pt}
    \caption{Ablation of the number of retained ranked targets $R$ in SRC distillation. We report mIoU (\%) on ShapeNetPart.}
    \label{tab:ablation_topm}
    \begin{tabular}{lcccccc}
    \toprule
    Top-$R$ & 1 & 2 & 3 & 4 & 5 & all \\
    \midrule
    mIoU$\uparrow$ & 57.9 & 58.3 & 58.6 & \textbf{59.0} & 58.7 & 58.6 \\
    \bottomrule
    \end{tabular}
\end{table}

\subsubsection{SRC marginal constraints}
\label{subsubsec:ablation_src}

We further analyze the internal design of SRC distillation by comparing different target distributions. SRC differs from standard prototype distillation in two key aspects: Sinkhorn assignment and long-tail-aware target marginal constraints.
Tab.~\ref{tab:ablation_src} shows that the softmax prototype target achieves 57.6\% mIoU but lacks global optimal transport constraints, making it more susceptible to unbalanced assignments. A uniform Sinkhorn target introduces transport constraints and improves performance to 58.6\% mIoU. However, a rigid uniform marginal may over-allocate probability mass to rare or unsupported prototypes in long-tailed 3D scenes. The long-tail Sinkhorn target adapts the prototype marginal according to dataset-level prototype frequencies. Combined with top-$R$ ranked sparsification, the full SRC configuration achieves the best performance of 59.0\% mIoU.

\begin{table}[!t]
    \centering
    \tablefont
    \renewcommand{\arraystretch}{\tablearraystretch}
    \setlength{\tabcolsep}{2.5pt}
    \caption{Ablation of SRC distillation on ShapeNetPart.}
    \label{tab:ablation_src}
    \begin{tabular}{@{}lcc@{}}
    \toprule
    SRC variant &  mIoU$\uparrow$ & Comment \\
    \midrule
    Softmax prototype target & 57.6 & no OT constraint \\
    Uniform Sinkhorn target & 58.6 & uniform marginal \\
    \rowcolor{rowcolor}Long-tail Sinkhorn target & \bf59.0 & adaptive marginal (full SRC) \\
    \bottomrule
    \end{tabular}
\end{table}

\subsubsection{Token serialization}
\label{subsubsec:ablation_serialization}

Since CLIP-ViT is pretrained on ordered 2D image patches, preserving spatial locality in the serialized 3D token sequence may improve its compatibility with irregular point clouds. We therefore compare random ordering, coordinate sorting, and Hilbert space-filling curves on ScanNetv2.
Tab.~\ref{tab:ablation_serialization} shows that random ordering achieves 56.5\% mIoU, while coordinate sorting slightly improves performance to 56.6\%. The Hilbert curve achieves the best result of 56.8\% mIoU, suggesting that locality-preserving serialization slightly better matches the spatial bias of the frozen transformer.

\begin{table}[!hbt]
    \centering
    \tablefont
    \renewcommand{\arraystretch}{\tablearraystretch}
    \setlength{\tabcolsep}{15pt}
    \caption{Ablation of token serialization strategies for annotation-free semantic segmentation on ScanNetv2. We report mIoU (\%), and $\Delta$ mIoU is measured against random ordering}
    \label{tab:ablation_serialization}
    \begin{tabular}{lcc}
    \toprule
    Serialization strategy & mIoU$\uparrow$ & $\Delta$ mIoU$\uparrow$\\
    \midrule
    Random ordering & 56.5 & --\\
    Coordinate sorting & 56.6 & +0.1\\
    \rowcolor{rowcolor}Hilbert curve (ours) & \textbf{56.8} & +0.3 \\
    \bottomrule
    \end{tabular}
\end{table}

\subsubsection{Superpoint-level dense prediction}
\label{subsubsec:ablation_dense_prediction}

We compare direct point-level projection with the proposed superpoint-level aggregation on joint training. This ablation evaluates how sparse token features should be mapped back to point-wise outputs.
Tab.~\ref{tab:ablation_dense_prediction} shows that superpoint-level aggregation consistently improves performance across ShapeNetPart, ScanNetv2, and S3DIS. Direct point-level projection is sensitive to noisy token assignments, whereas superpoint aggregation pools features over geometrically coherent regions, improving semantic consistency and local boundary preservation.

\begin{table}[!hbt]
    \centering
    \tablefont
    \renewcommand{\arraystretch}{\tablearraystretch}
    \setlength{\tabcolsep}{4pt}
    \caption{Ablation of dense prediction strategies. We report mIoU (\%) on annotation-free segmentation benchmarks.}
    \label{tab:ablation_dense_prediction}
    \begin{tabular}{lccc}
    \toprule
    Prediction strategy & ShapeNetPart & ScanNetv2 & S3DIS \\
    \midrule
    Point-level projection & 57.6 & 56.2 & 48.9 \\
    \rowcolor{rowcolor}Superpoint-level aggregation & \textbf{59.3} & \textbf{59.2} & \textbf{49.7} \\
    \bottomrule
    \end{tabular}
\end{table}

%% file: table/shapenet.tex
\begin{table*}[t]
\centering
\tablefont
\renewcommand{\arraystretch}{\tablearraystretch}
\setlength{\tabcolsep}{2.2pt}
\caption{Annotation-free part segmentation on ShapeNetPart~\cite{chang2015shapenet}. PointCLIPv2$^\ast$ denotes our reproduced result. ``--'' indicates unavailable per-category results.}
\label{tab:part_seg}
\resizebox{\textwidth}{!}{
\begin{tabular}{lccccccccccccccccc}
\toprule
Method & mIoU$\uparrow$
& Airplane & Bag & Cap & Car & Chair & Earphone & Guitar & Knife
& Lamp & Laptop & Motorbike & Mug & Pistol & Rocket & Skateboard & Table \\
\midrule
\multicolumn{18}{c}{\textit{Features extracted from multi-view projections using VLMs}} \\
\midrule
PointCLIP~\cite{zhang2022pointclip}
& 24.8
& 22.0 & 44.8 & 13.4 & -- & 18.7 & 28.3 & 22.7 & 24.8
& -- & 22.9 & -- & 48.6 & -- & 22.7 & 42.7 & 45.5 \\

PointCLIPv2~\cite{zhu2023pointclip}
& 44.3
& 33.5 & 60.4 & 52.8 & -- & 51.5 & 56.5 & 71.5 & 66.7
& -- & 61.6 & -- & 48.0 & -- & 49.6 & 43.9 & 61.1 \\

PointCLIPv2$^\ast$~\cite{zhu2023pointclip}
& 51.8
& 33.5 & 60.4 & 52.9 & 27.2 & 51.5 & 56.5 & 71.5 & 76.7
& 44.7 & 61.5 & 31.5 & 48.0 & 46.1 & \textbf{49.6} & 43.9 & 61.1 \\

GeoZe~\cite{mei2024geometrically}
& 57.4
& 33.4 & 69.4 & \textbf{59.7} & 34.8 & 66.2 & \textbf{66.9} & 72.0 & 78.3
& \textbf{45.2} & 74.7 & 28.1 & \textbf{61.8} & 48.2 & 47.2 & \textbf{45.8} & 63.9 \\

\midrule
\multicolumn{18}{c}{\textit{Features extracted from Point-M2AE~\cite{zhang2022point} fully distilled from 2D VLMs}} \\
\midrule
PartDistill~\cite{umam2024partdistill}
& 53.8
& 37.5 & 62.6 & 55.5 & -- & 56.4 & 55.6 & 71.7 & 76.9
& -- & 67.4 & -- & 53.5 & -- & -- & -- & 62.9 \\

\midrule
\multicolumn{18}{c}{\textit{Features extracted from a frozen CLIP-ViT-B/16 with a jointly trained 3D tokenizer}} \\
\midrule
\rowcolor{rowcolor}
\ourmethod (ours)
& \textbf{59.3}
& \textbf{38.6} & \textbf{73.3} & 53.7 & \textbf{38.3}
& \textbf{70.7} & 66.3 & \textbf{76.3} & \textbf{79.0}
& 28.1 & \textbf{80.2} & 22.8 & 54.4 & \textbf{54.3}
& 44.9 & 43.9 & \textbf{68.6} \\
\bottomrule
\end{tabular}
}
\end{table*}

%% file: fig_tex/partseg.tex

\newcommand{\partsegscw}{.21\columnwidth}

\begin{figure}[t]
    \centering
    \setlength{\tabcolsep}{0.4pt}

    \begin{tabular}{@{}cccc@{}}
    \begin{overpic}[width=\partsegscw]{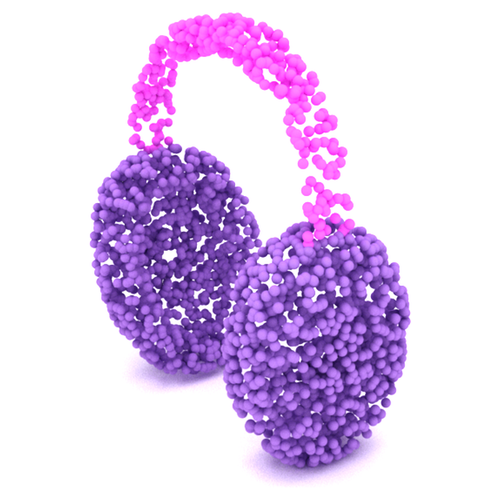}
        \put(-8,38){\rotatebox{90}{\color{black}\scriptsize GT}}
    \end{overpic} &
    \begin{overpic}[width=\partsegscw]{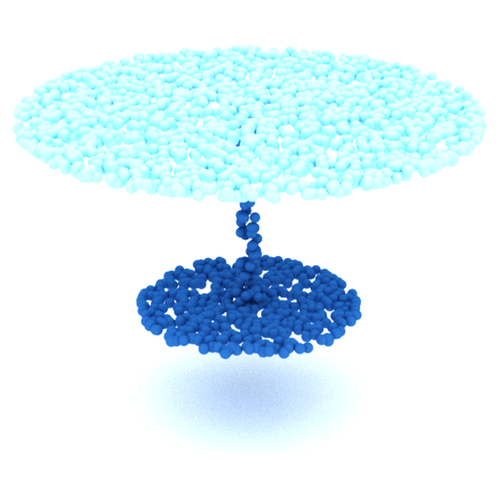}
    \end{overpic} &
    \begin{overpic}[width=\partsegscw]{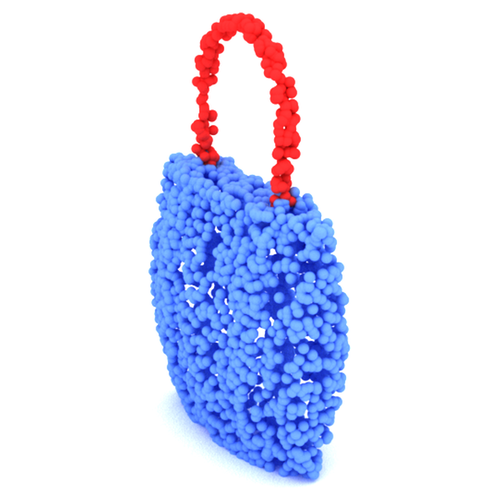}
    \end{overpic} &
    \begin{overpic}[width=\partsegscw]{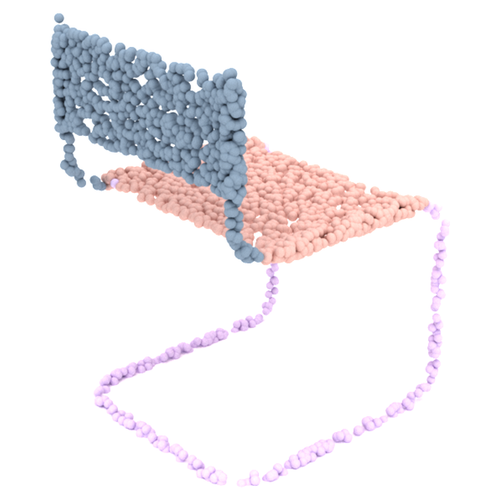}
    \end{overpic} \\[-1.2mm]

    \begin{overpic}[width=\partsegscw]{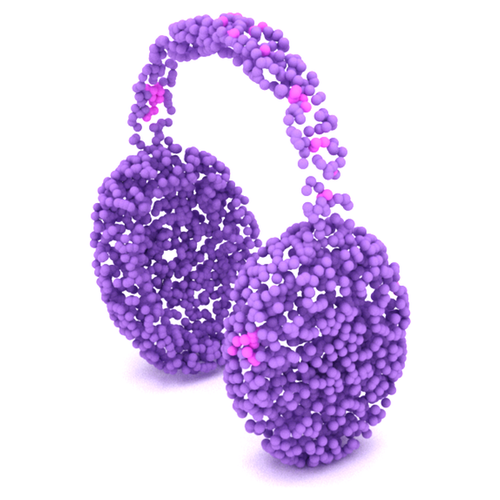}
        \put(-8,6){\rotatebox{90}{\color{black}\scriptsize PointCLIPv2}}
    \end{overpic} &
    \begin{overpic}[width=\partsegscw]{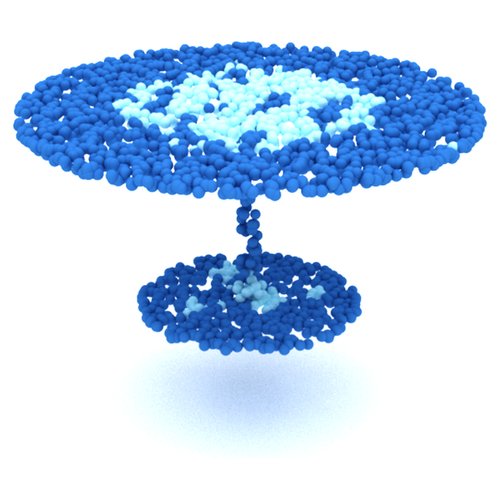}
    \end{overpic} &
    \begin{overpic}[width=\partsegscw]{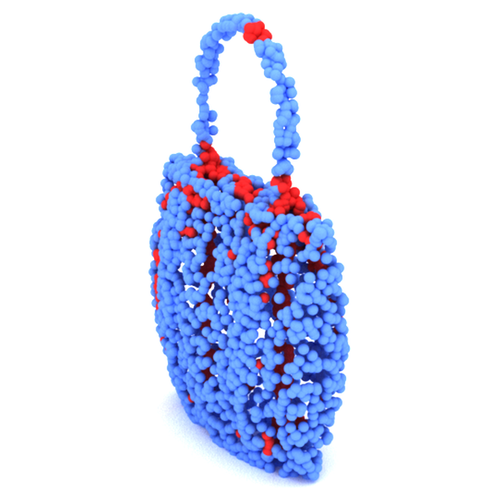}
    \end{overpic} &
    \begin{overpic}[width=\partsegscw]{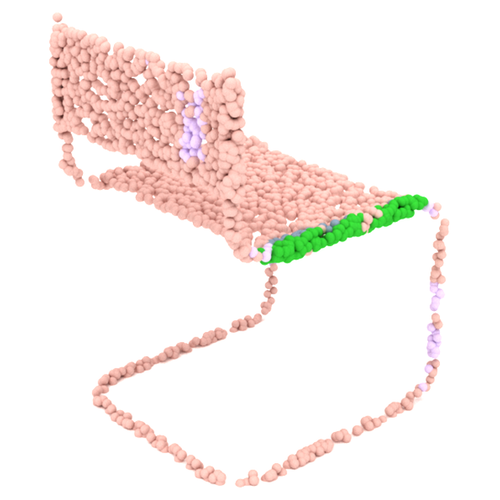}
    \end{overpic} \\[-1.2mm]

    \begin{overpic}[width=\partsegscw]{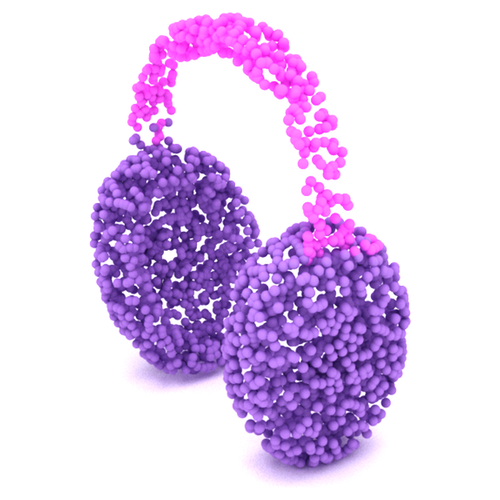}
        \put(-8,27){\rotatebox{90}{\color{black}\scriptsize GeoZe}}
    \end{overpic} &
    \begin{overpic}[width=\partsegscw]{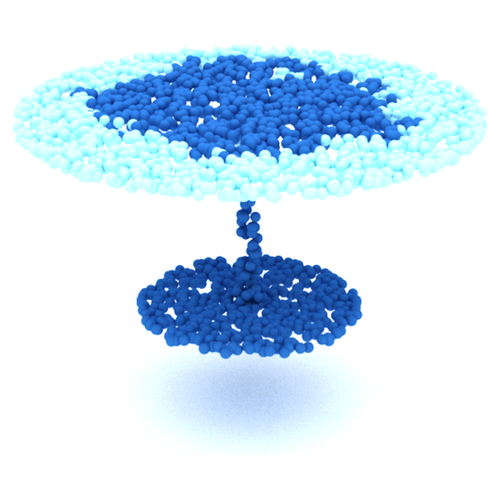}
    \end{overpic} &
    \begin{overpic}[width=\partsegscw]{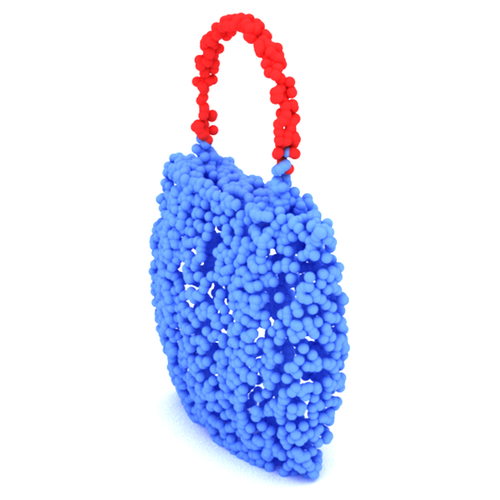}
    \end{overpic} &
    \begin{overpic}[width=\partsegscw]{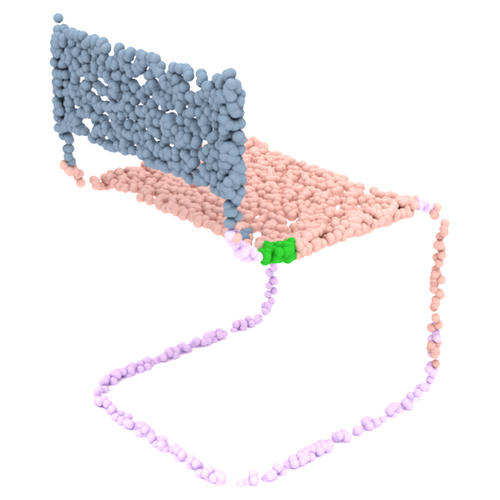}
    \end{overpic} \\[-1.2mm]

    \begin{overpic}[width=\partsegscw]{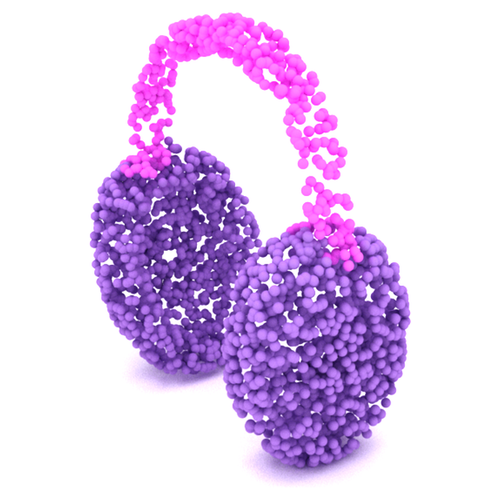}
        \put(-8,18){\rotatebox{90}{\color{black}\scriptsize \ourmethod}}
    \end{overpic} &
    \begin{overpic}[width=\partsegscw]{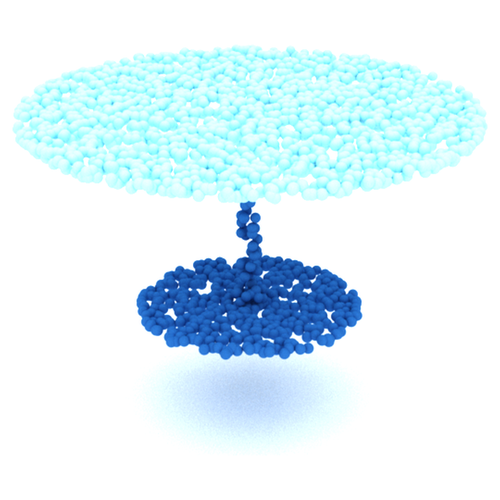}
    \end{overpic} &
    \begin{overpic}[width=\partsegscw]{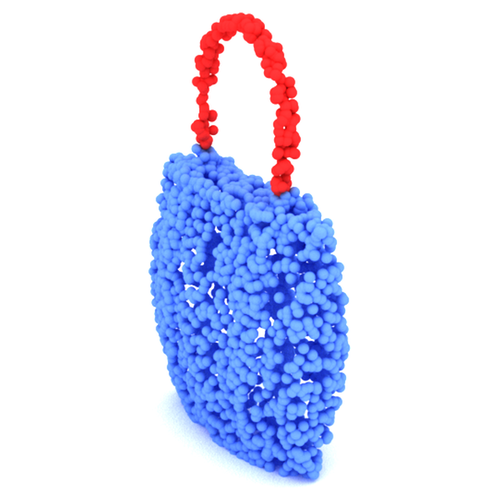}
    \end{overpic} &
    \begin{overpic}[width=\partsegscw]{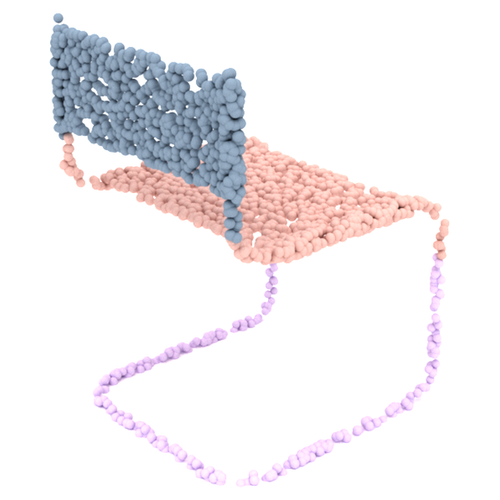}
    \end{overpic} \\[-0.8mm]

    {\scriptsize earphone} &
    {\scriptsize table} &
    {\scriptsize bag} &
    {\scriptsize chair} \\

    \end{tabular}
    \vspace{-2mm}
    \caption{Qualitative part segmentation results on ShapeNetPart. Rows show ground truth and predictions from PointCLIPv2, GeoZe, and \ourmethod. Superpoint visualizations are omitted for compactness.}
    \label{fig:partseg}
    \vspace{-2mm}
\end{figure}

%% file: table/indoor.tex
\begin{table*}[t]
  \centering
  \tablefont
  \renewcommand{\arraystretch}{\tablearraystretch}
  \setlength{\tabcolsep}{4pt}
  \caption{Annotation-free semantic segmentation on ScanNetV2 and S3DIS. --'' indicates unavailable results. Bold denotes the best annotation-free mIoU. Supervised baselines are upper-reference results. $^\dagger$ denotes reproduced results. $^\ddagger$ denotes clustering methods evaluated with Hungarian label alignment; their S3DIS results follow the 12-class protocol excluding clutter.}
  \label{tab:clip_sem_seg}
  \begin{tabular}{lccccccccc}
    \toprule
    \multirow{2}{*}{Method}
    & \multirow{2}{*}{Venue}
    & \multirow{2}{*}{Input}
    & \multirow{2}{*}{Backbone / Rep.}
    & \multirow{2}{*}{\#Params}
    & \multirow{2}{*}{Semantic}
    & \multicolumn{2}{c}{ScanNetv2 (val)}
    & \multicolumn{2}{c}{S3DIS (Area 5)} \\
    \cmidrule(lr){7-8} \cmidrule(lr){9-10}
    & & & & & & mIoU$\uparrow$ & mAcc$\uparrow$ & mIoU$\uparrow$ & mAcc$\uparrow$ \\
    \midrule

    \multicolumn{10}{c}{\textit{Features extracted from 3D backbones fully trained using labeled data}} \\
    \midrule
    Scratch
    & --
    & pcd
    & MinkowskiNet18A~\cite{choy20194d}
    & 15.5M
    & GT labels
    & 65.6$^\dagger$ & 73.9$^\dagger$ & 65.4$^\dagger$ & 71.7$^\dagger$ \\

    Scratch
    & --
    & pcd
    & PViT~\cite{qian2024pix4point}
    & 27.1M
    & GT labels
    & 60.1$^\dagger$ & 67.6$^\dagger$ & 64.4 & 69.9 \\

    \midrule
    \multicolumn{10}{c}{\textit{Features extracted from multi-view projections using VLMs}} \\
    \midrule
    OpenScene~\cite{peng2023openscene}
    & CVPR 2023
    & img
    & 2D projection
    & 0M
    & LSeg~\cite{li2022languagedriven}
    & 50.0 & 62.7 & 40.3$^\dagger$ & 37.5$^\dagger$ \\

    GeoZe~\cite{mei2024geometrically}
    & CVPR 2024
    & img
    & 2D projection
    & 0M
    & LSeg~\cite{li2022languagedriven}
    & 54.7 & 69.3 & 41.4$^\dagger$ & 51.1$^\dagger$ \\

    GeoZe~\cite{mei2024geometrically}
    & CVPR 2024
    & img
    & 2D projection
    & 0M
    & OpenSeg~\cite{ghiasi2022scaling}
    & 47.8 & 71.4 & 30.2$^\dagger$ & 54.7$^\dagger$ \\

    \midrule
    \multicolumn{10}{c}{\textit{Unsupervised 3D segmentation with clustering-based label alignment}} \\
    \midrule
    LogoSP$^\ddagger$~\cite{zhang2025logosp}
    & CVPR 2025
    & pcd+img
    & SparseConv~\cite{graham20183d}
    & --
    & DINOv2~\cite{oquab2023dinov2}
    & 35.8 & 50.8 & 46.5 & 55.9 \\

    GrowSP++$^\ddagger$~\cite{zhang2026growspplusplus}
    & TPAMI 2026
    & pcd+img
    & SparseConv~\cite{graham20183d}
    & --
    & DINOv2~\cite{oquab2023dinov2}
    & 33.2 & 48.8 & 46.6 & 60.1 \\

    P-SLCR$^\ddagger$~\cite{zhan2026pslcr}
    & AAAI 2026
    & pcd
    & SparseConv~\cite{graham20183d}
    & --
    & clusters
    & 29.0 & 49.0 & 47.1 & 57.2 \\

    PointGS$^\ddagger$~\cite{song2026pointgs}
    & CVPR 2026
    & pcd+img
    & 3DGS~\cite{kerbl20233d}/SAM~\cite{kirillov2023segment}
    & --
    & SAM~\cite{kirillov2023segment}
    & 36.7 & -- & 49.3 & 66.1 \\

    \midrule
    \multicolumn{10}{c}{\textit{Features extracted from 3D backbones fully distilled from 2D VLMs}} \\
    \midrule
    CLIP-FO3D~\cite{zhang2023clip}
    & imgCCVW 2023
    & pcd
    & MinkowskiNet16~\cite{choy20194d}
    & --
    & CLIP~\cite{radford2021clip}
    & 30.2 & 49.1 & 22.3 & 32.8 \\

    OpenScene~\cite{peng2023openscene}
    & CVPR 2023
    & pcd
    & MinkowskiNet18A~\cite{choy20194d}
    & 15.5M
    & LSeg~\cite{li2022languagedriven}
    & 54.2 & 66.6 & -- & -- \\

    PGOV3D~\cite{zhang2025pgov3d}
    & ACM MM 2025
    & pcd
    & SparseConv~\cite{graham20183d}
    & --
    & CLIP~\cite{radford2021clip}
    & 59.5 & 73.2 & -- & -- \\

    Mosaic3D~\cite{lee2025mosaic3d}
    & CVPR 2025
    & pcd
    & MinkowskiNet34C~\cite{choy20194d}
    & 43.7M
    & Recap-CLIP~\cite{li2024recaption}
    & 42.5 & 66.5 & 27.2 & 43.3 \\

    GeoGuide~\cite{tao2026geoguide}
    & CVPR 2026
    & pcd
    & Pretrained PTv3~\cite{wu2024point,wu2025sonata}
    & --
    & LSeg~\cite{li2022languagedriven}
    & \textbf{59.8} & 72.5 & -- & -- \\

    \midrule
    \multicolumn{10}{c}{\textit{Features extracted from a frozen CLIP-ViT-B/16 with a jointly trained 3D tokenizer}} \\
    \midrule
    \rowcolor{rowcolor}
    \ourmethod (ours)
    & Ours
    & pcd
    & Frozen ViT-B/16~\cite{radford2021clip}
    & 3.39M
    & CLIP~\cite{radford2021clip}
    & 59.2 & 72.4 & \textbf{49.7} & 61.1 \\
    \bottomrule
  \end{tabular}
  \vspace{-2mm}
\end{table*}

%% file: fig_tex/indoor.tex

\begin{figure*}[t]
    \centering
    \def\swmerge{0.15\textwidth} 
    \setlength{\tabcolsep}{1pt}
    \renewcommand{\arraystretch}{0.5}
    \begin{tabular}{@{}ccc@{\hskip 1.5mm};{0.8pt/2pt \color{blue}}@{\hskip 1.5mm}ccc@{}}
    \multicolumn{3}{c}{\footnotesize ScanNetv2} &
    \multicolumn{3}{;{0.8pt/2pt \color{black}}c}{\footnotesize S3DIS} \\[0.3mm]

    \begin{overpic}[width=\swmerge]{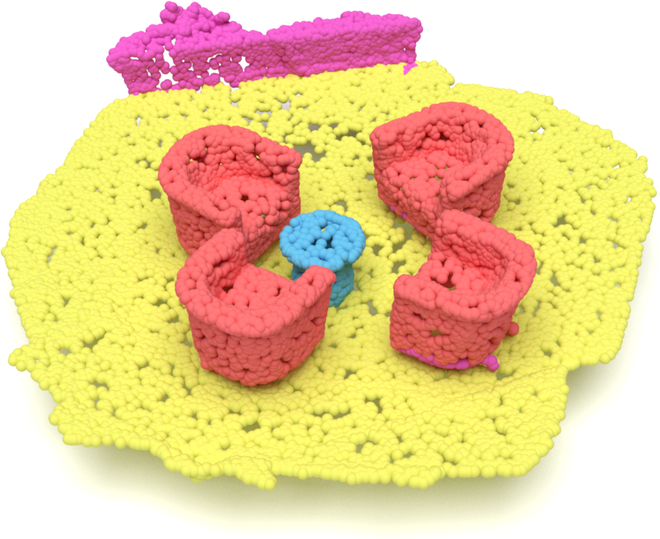}
        \put(-10,42){\rotatebox{90}{\color{black}\footnotesize GT}}
    \end{overpic} &
    \begin{overpic}[width=\swmerge]{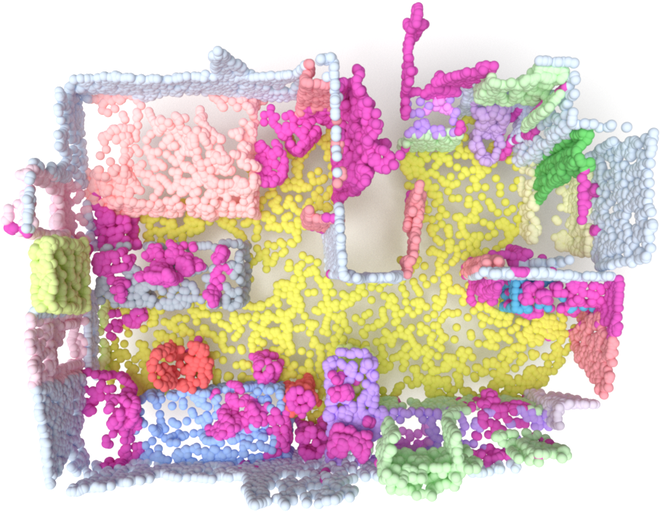}
    \end{overpic} &
    \begin{overpic}[width=\swmerge]{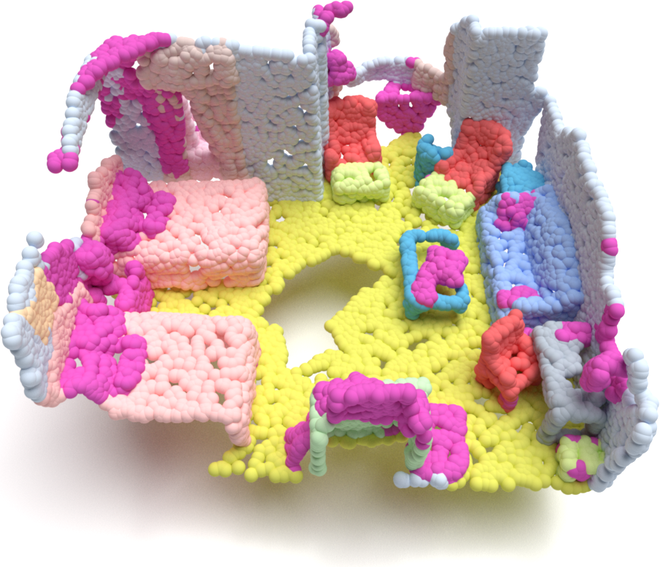}
    \end{overpic} &
    \begin{overpic}[width=\swmerge]{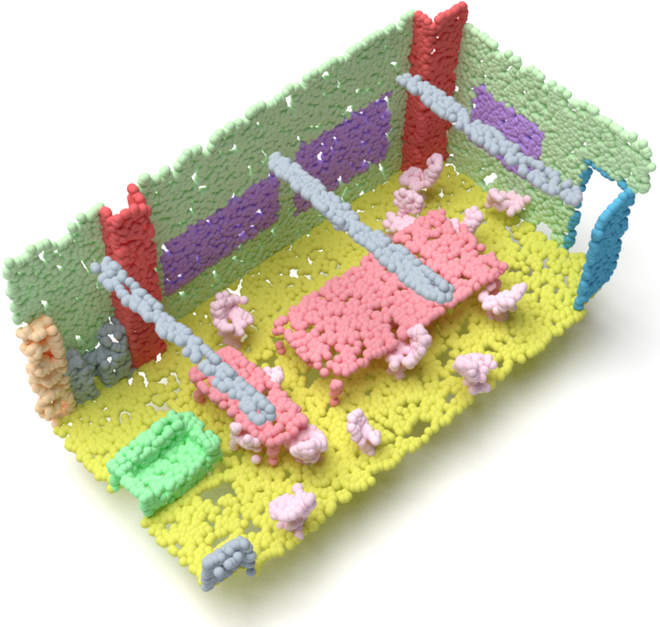}
    \end{overpic} &
    \begin{overpic}[width=\swmerge]{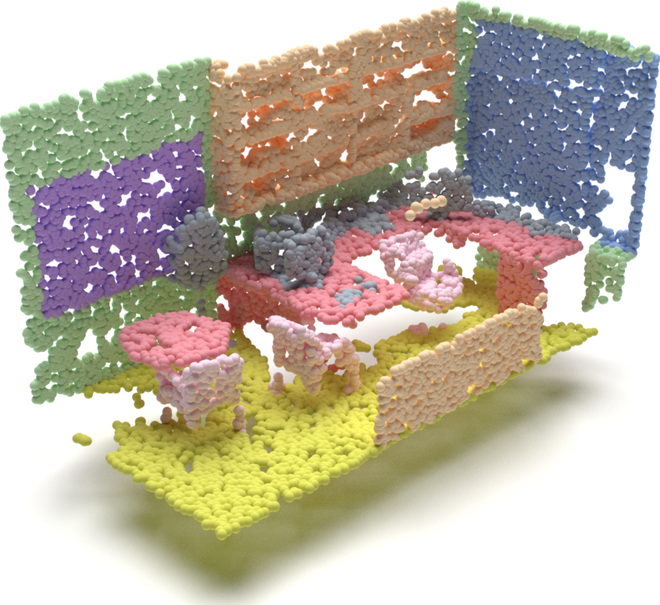}
    \end{overpic} &
    \begin{overpic}[width=\swmerge]{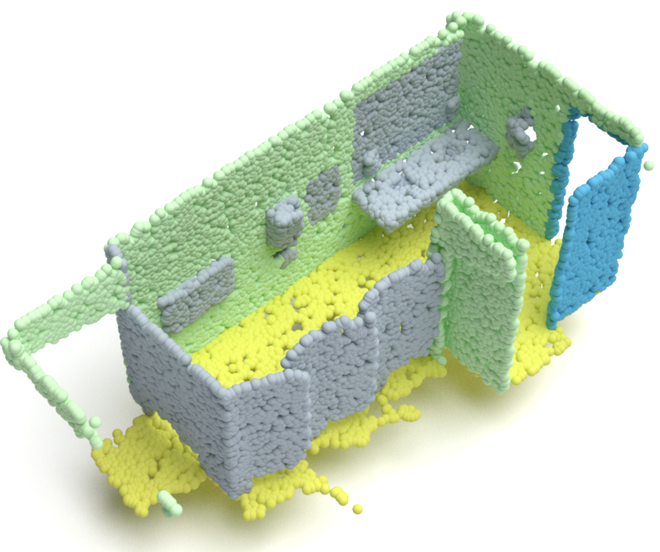}
    \end{overpic} \\[-1.5mm]

    \begin{overpic}[width=\swmerge]{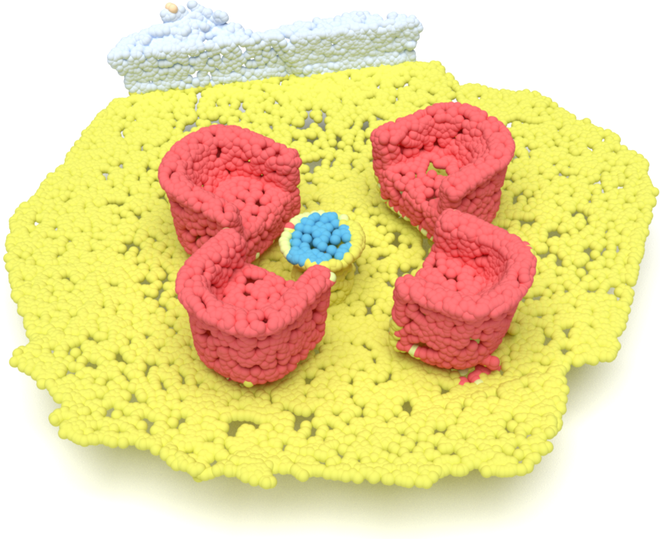}
        \put(-10,34){\rotatebox{90}{\color{black}\footnotesize OpenScene}}
    \end{overpic} &
    \begin{overpic}[width=\swmerge]{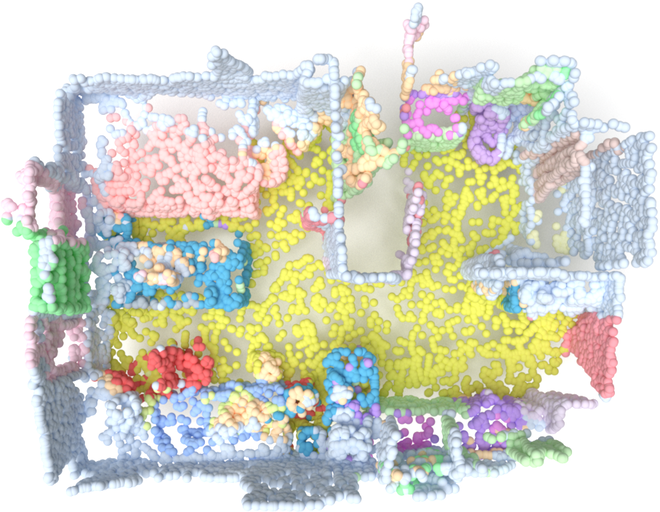}
    \end{overpic} &
    \begin{overpic}[width=\swmerge]{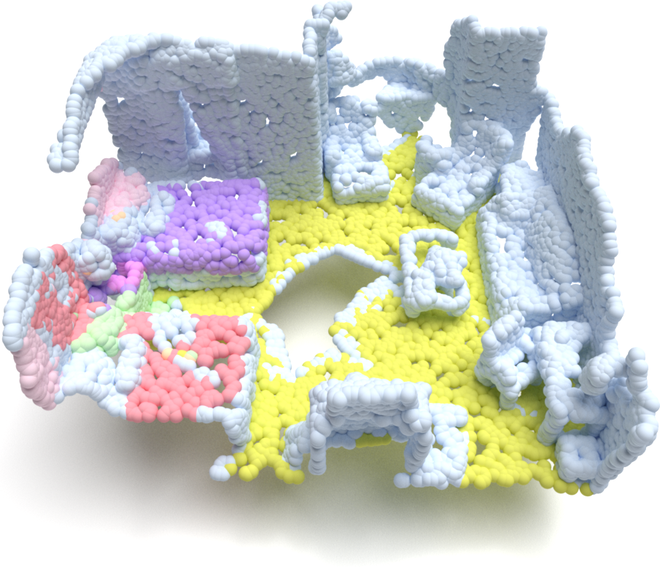}
    \end{overpic} &
    \begin{overpic}[width=\swmerge]{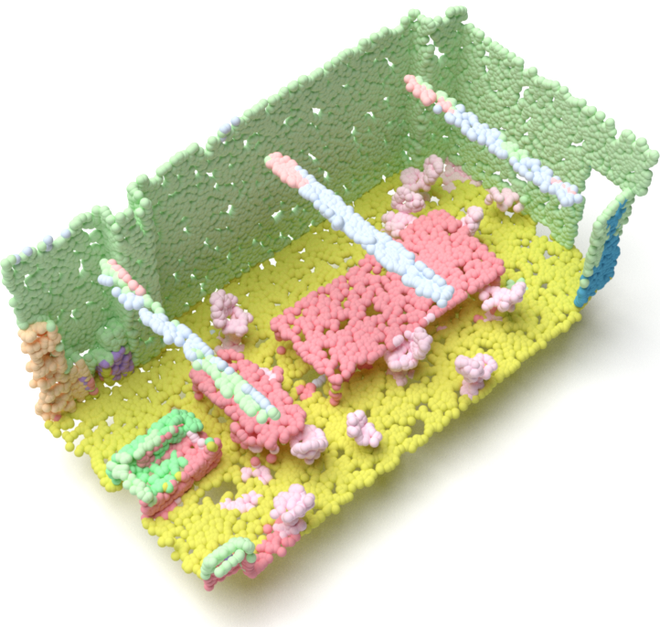}
    \end{overpic} &
    \begin{overpic}[width=\swmerge]{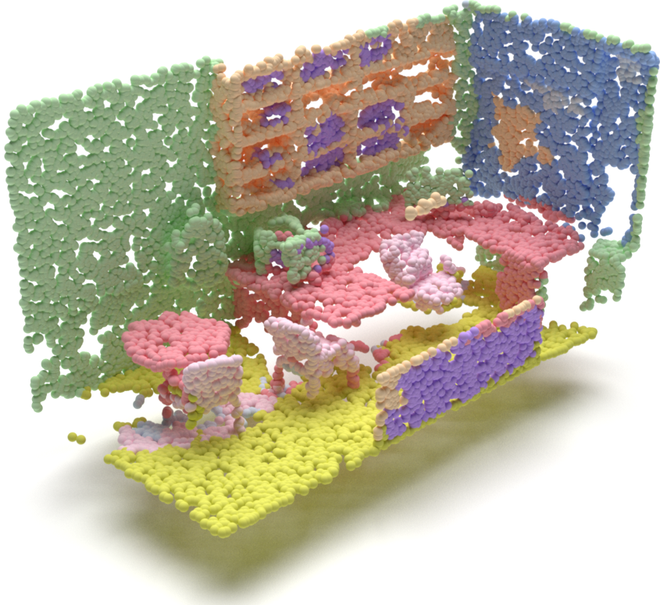}
    \end{overpic} &
    \begin{overpic}[width=\swmerge]{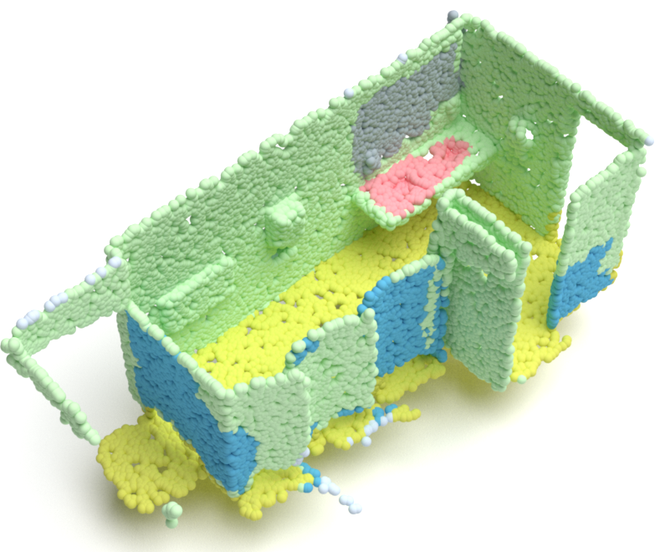}
    \end{overpic} \\[-1.5mm]

    \begin{overpic}[width=\swmerge]{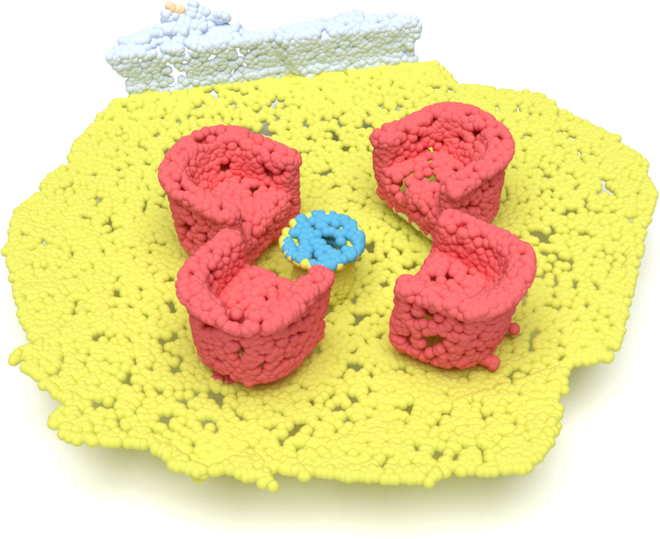}
        \put(-10,40){\rotatebox{90}{\color{black}\footnotesize GeoZe}}
    \end{overpic} &
    \begin{overpic}[width=\swmerge]{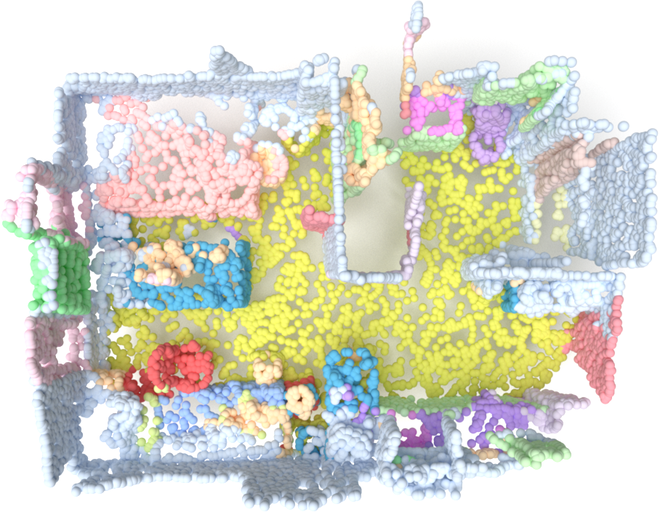}
    \end{overpic} &
    \begin{overpic}[width=\swmerge]{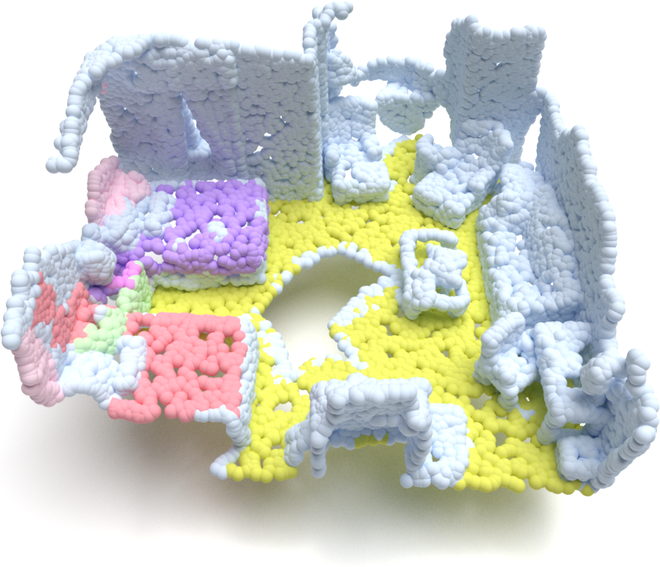}
    \end{overpic} &
    \begin{overpic}[width=\swmerge]{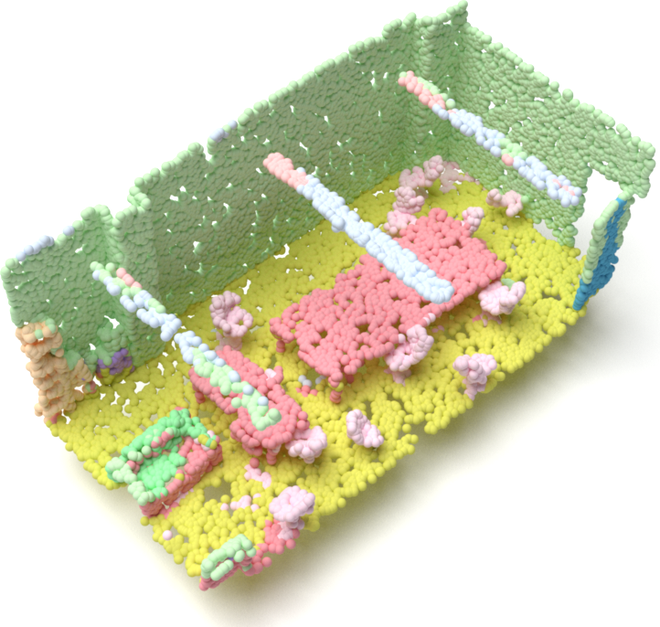}
    \end{overpic} &
    \begin{overpic}[width=\swmerge]{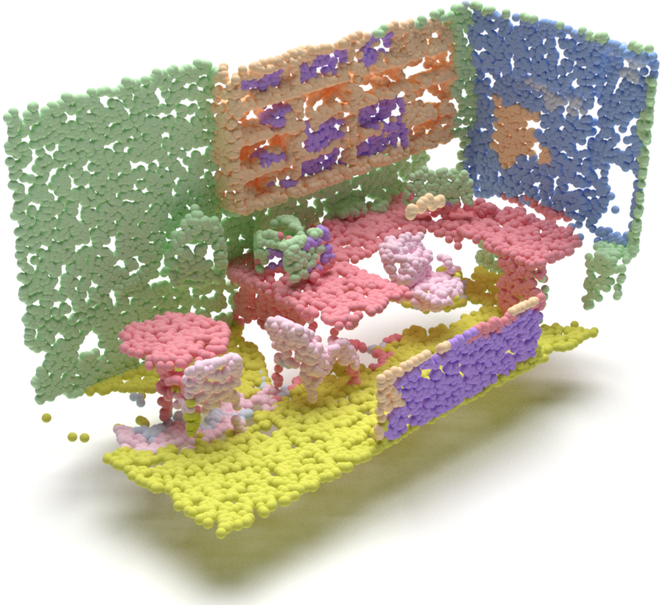}
    \end{overpic} &
    \begin{overpic}[width=\swmerge]{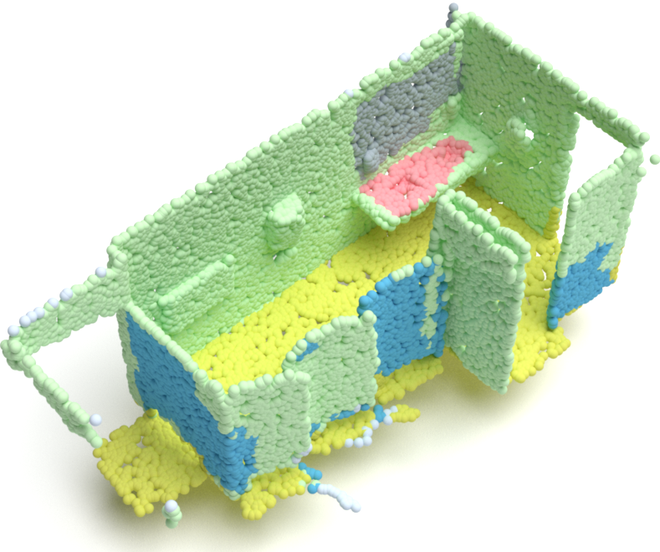}
    \end{overpic} \\[-1.5mm]

    \begin{overpic}[width=\swmerge]{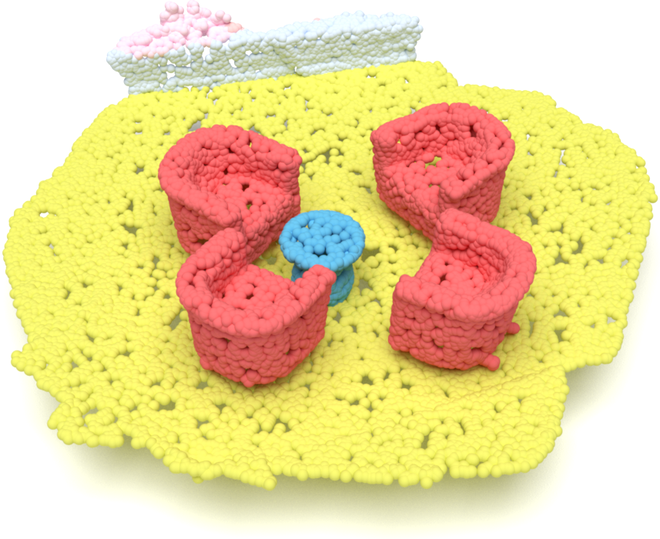}
        \put(-10,38){\rotatebox{90}{\color{black}\footnotesize \ourmethod}}
    \end{overpic} &
    \begin{overpic}[width=\swmerge]{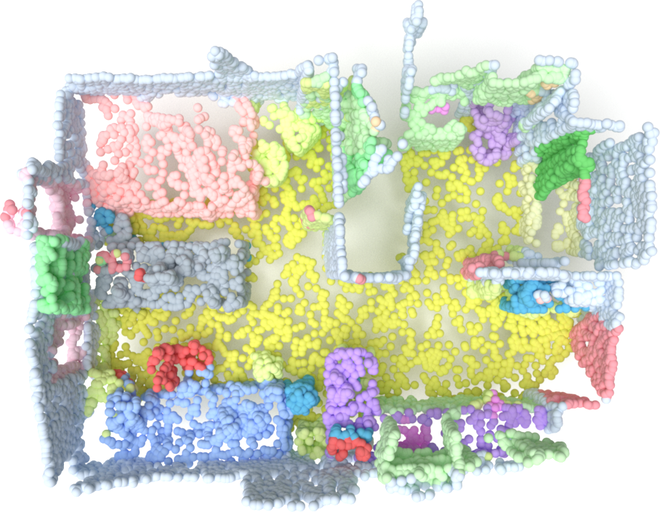}
    \end{overpic} &
    \begin{overpic}[width=\swmerge]{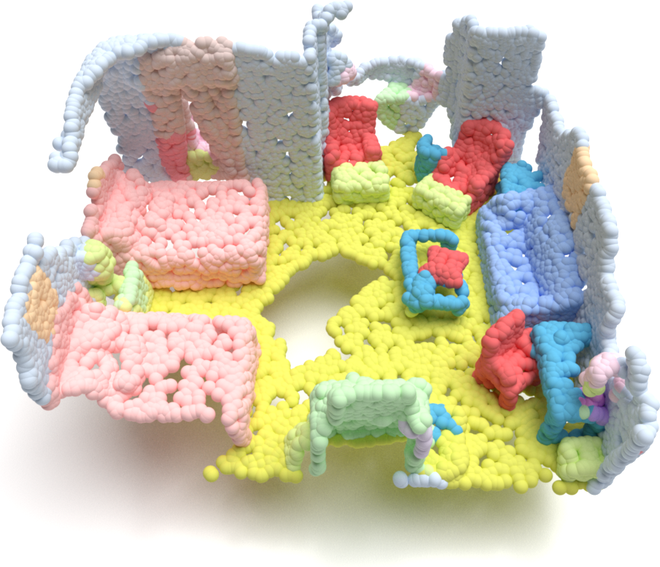}
    \end{overpic} &
    \begin{overpic}[width=\swmerge]{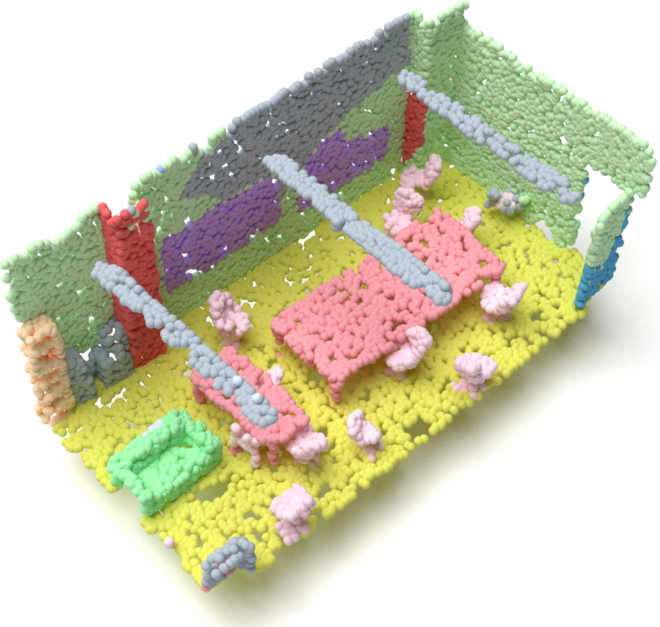}
    \end{overpic} &
    \begin{overpic}[width=\swmerge]{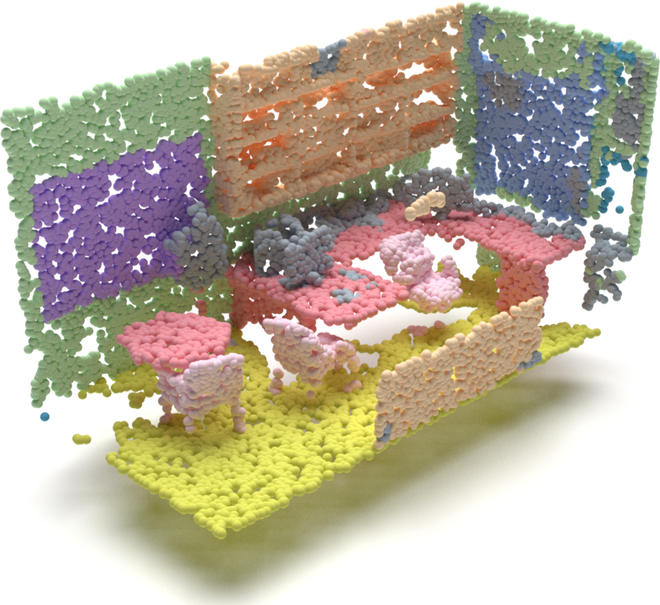}
    \end{overpic} &
    \begin{overpic}[width=\swmerge]{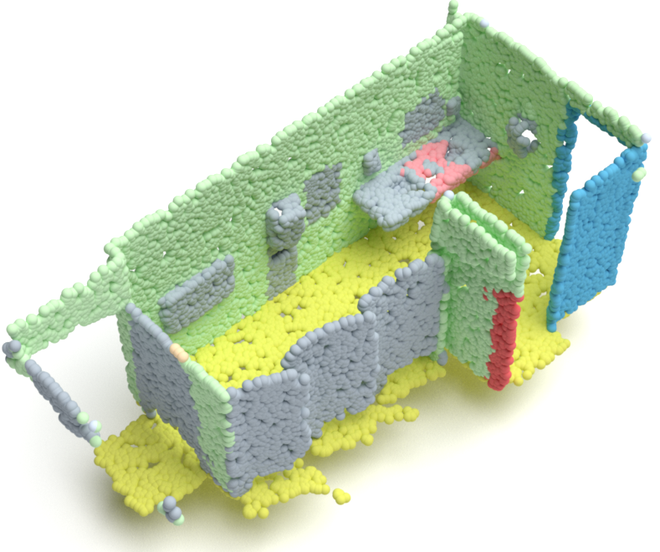}
    \end{overpic} \\[-1.5mm]

    \end{tabular}
    \caption{Qualitative comparison of open-vocabulary semantic segmentation on ScanNetv2~\cite{dai2017scannet} (left three columns) and S3DIS~\cite{armeni_cvpr16} (right three columns), separated by the dotted line. Rows show ground truth and predictions from OpenScene~\cite{jiang2024open}, GeoZe~\cite{mei2024geometrically}, and our \ourmethod.}
    \label{fig:scannet_s3dis}
\end{figure*}

%% file: fig_tex/kitti.tex
\newcommand{\kittiwidth}{.32\columnwidth}

\begin{figure}[t]
    \centering
    \setlength{\tabcolsep}{0pt}
    \renewcommand{\arraystretch}{0.5}
    \begin{tabular}{@{}c@{}c@{}c@{}}
    \begin{overpic}[width=\kittiwidth]{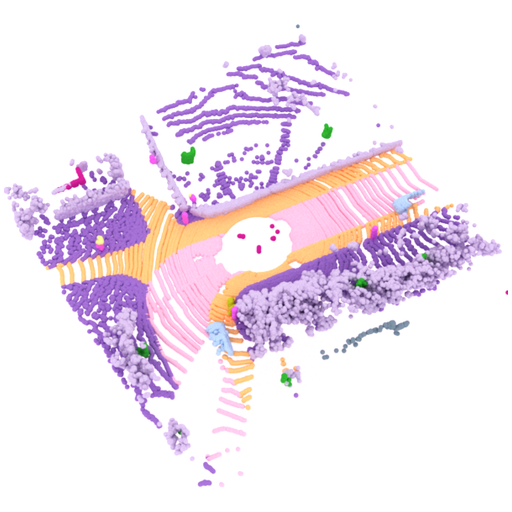}
        \put(-10,45){\rotatebox{90}{\color{black}\footnotesize GT}}
    \end{overpic} &
    \begin{overpic}[width=\kittiwidth]{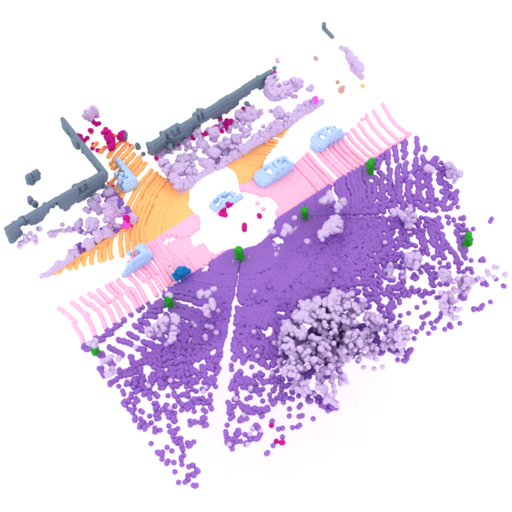}
    \end{overpic} &
    \begin{overpic}[width=\kittiwidth]{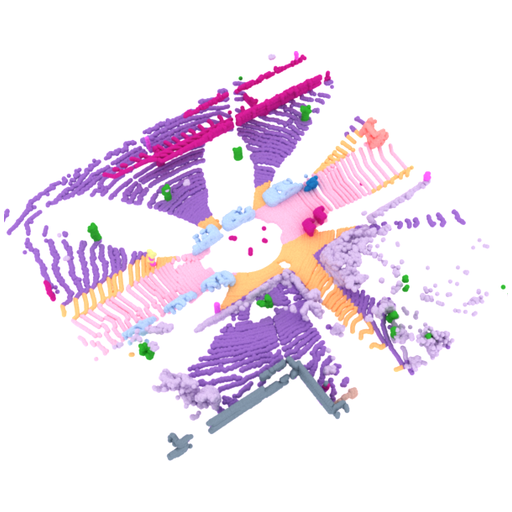}
    \end{overpic} \\[-1mm]
    
    \begin{overpic}[width=\kittiwidth]{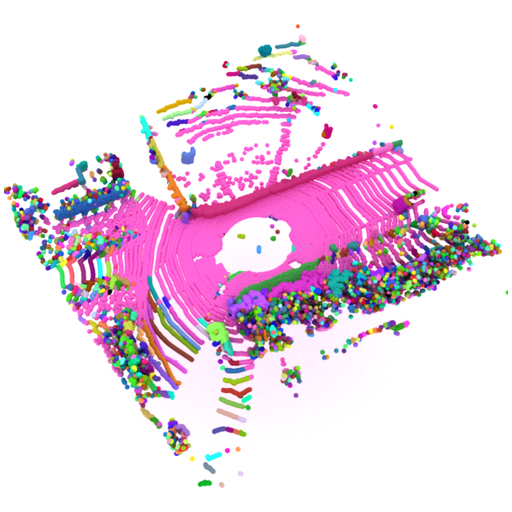}
        \put(-10,20){\rotatebox{90}{\color{black}\footnotesize Superpoints}}
    \end{overpic} &
    \begin{overpic}[width=\kittiwidth]{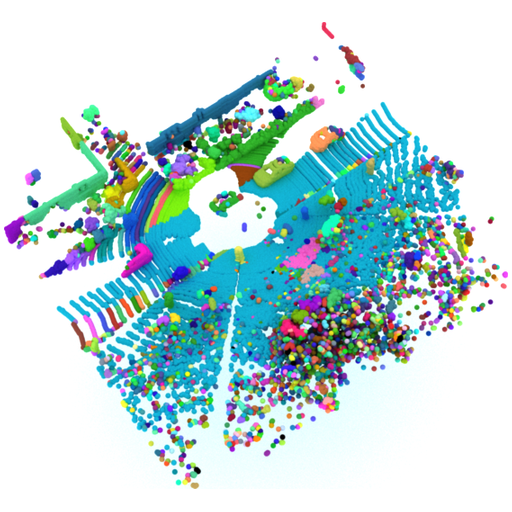}
    \end{overpic} &
    \begin{overpic}[width=\kittiwidth]{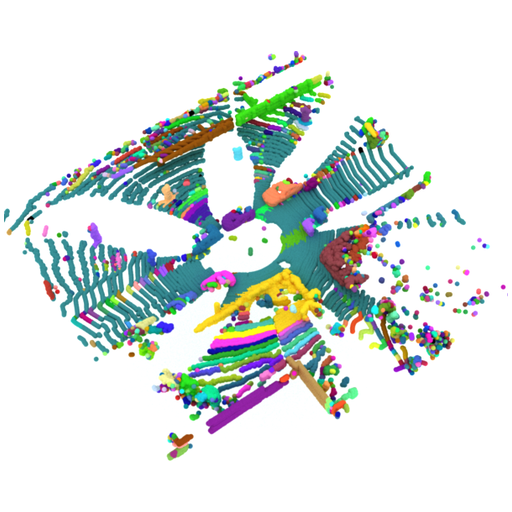}
    \end{overpic} \\[-1mm]
    
    \begin{overpic}[width=\kittiwidth]{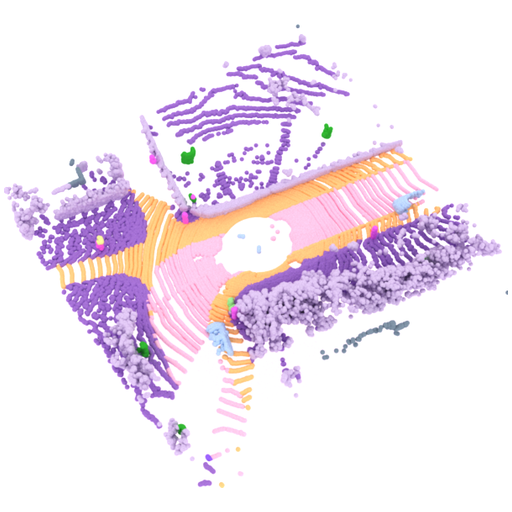}
        \put(-10,30){\rotatebox{90}{\color{black}\footnotesize \ourmethod}}
    \end{overpic} &
    \begin{overpic}[width=\kittiwidth]{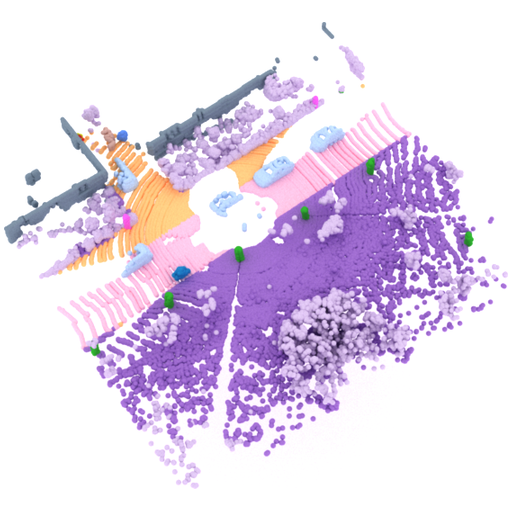}
    \end{overpic} &
    \begin{overpic}[width=\kittiwidth]{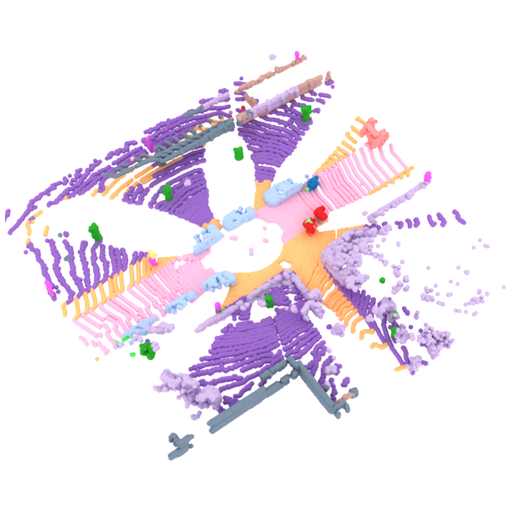}
    \end{overpic} \\[-1mm]
    \end{tabular}
    \vspace{-2mm}
    \caption{Qualitative semantic segmentation results on the SemanticKITTI~\cite{behley2019semantickitti} validation set (sequence-08). We compare \ourmethod predictions (bottom row) with superpoint over-segmentation tracks (middle row) and the corresponding ground-truth annotations (top row).}
    \label{fig:kitti}
    \vspace{-3mm}
\end{figure}

%% file: secs/5_cons.tex
\section{Conclusion}\label{sec:conclusion}
We presented \ourmethod, a parameter-efficient framework for adapting frozen 2D vision--language models to point clouds through unified scale-normalized tokenization. The proposed tokenizer estimates an input-adaptive geometric scale and uses it consistently for sparse voxelization, coordinate normalization, token-center construction, positional encoding, and Hilbert-based serialization. This design provides a stable 3D token interface across object-level shapes, indoor scenes, and outdoor LiDAR scans while preserving the pretrained CLIP backbone.
The tokenizer is trained without 3D semantic annotations through cross-modal distillation from multi-view foundation-model features. Local superpoint-level alignment transfers fine-grained 2D--3D correspondence, while Sinkhorn Ranked Contrastive distillation captures global semantic structure and mitigates collapse under imbalanced pseudo-labels. For dense prediction, superpoint-level aggregation and propagation bridge sparse token features with point-wise outputs.
Experiments demonstrate that \ourmethod achieves strong annotation-free segmentation and cross-domain generalization with substantially fewer trainable parameters than fully trained 3D backbones. These results suggest that scale-consistent 3D tokenization provides a promising interface for reusing frozen image foundation models on heterogeneous point-cloud data.

%% file: secs/profile.tex

\begin{IEEEbiography}[{\includegraphics[width=1in,height=1.25in,clip,keepaspectratio]{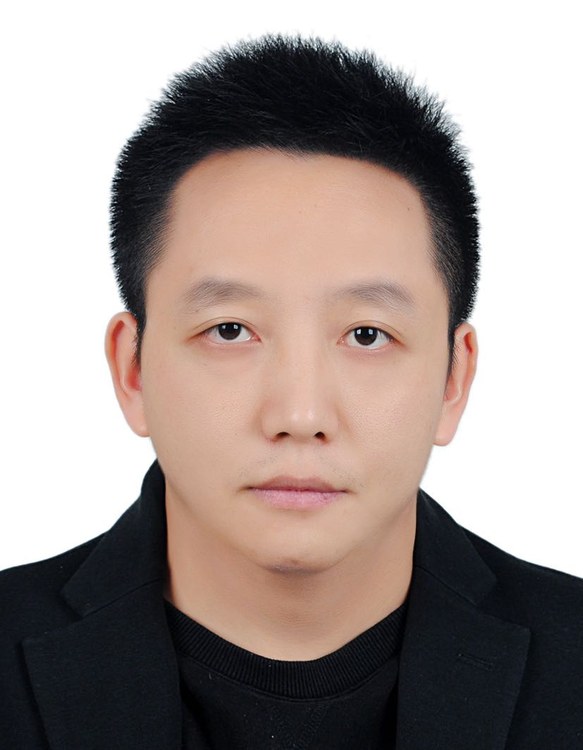}}]{Guofeng Mei}
received the Ph.D. degree in Software Engineering from the University of Technology Sydney (UTS), Sydney, Australia, in 2023. He is currently a Researcher at Fondazione Bruno Kessler, Trento, Italy. His research interests include 3D point cloud registration and matching, 3D vision understanding, natural language processing, vision-language models, machine learning, and optimization.
\end{IEEEbiography}

\begin{IEEEbiography}[{\includegraphics[width=1in,height=1.25in,clip,keepaspectratio]{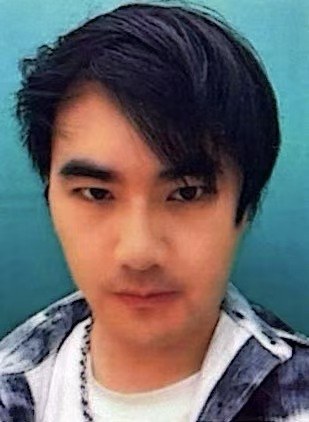}}]{Qinfeng Xiao}
is currently a Ph.D. student at The Hong Kong Polytechnic University, Hong Kong. He received the M.S. degree from Beijing Jiaotong University, China. His research interests include 3D shape matching, 3D representation learning, and 3D reconstruction.
\end{IEEEbiography}

\begin{IEEEbiography}[{\includegraphics[width=1in,height=1.25in,clip,keepaspectratio]{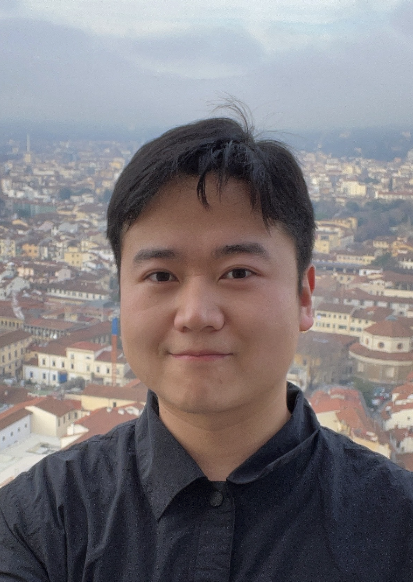}}]{Bin Ren}
is currently a Postdoctoral Associate at MBZUAI, Abu Dhabi, UAE. He received the Ph.D. degree in Artificial Intelligence from the Italian National Ph.D. Program in Artificial Intelligence, co-organized by the University of Pisa and the University of Trento, Italy. He received the M.S. degree from Peking University, China, and the B.Eng. degree from Central South University, China. His research interests lie at the intersection of computer vision and deep learning.
\end{IEEEbiography}

\begin{IEEEbiography}[{\includegraphics[width=1in,height=1.25in,clip,keepaspectratio]{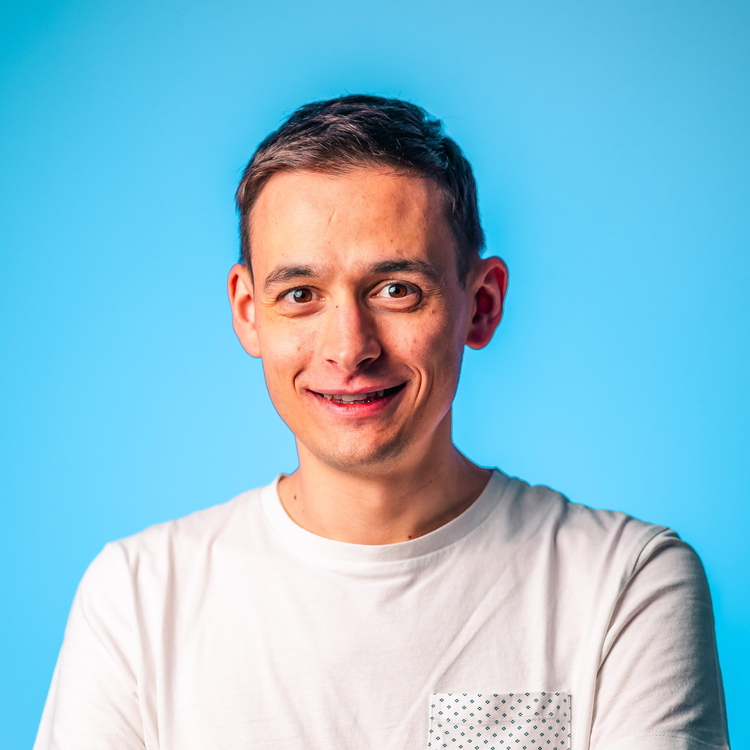}}]{Luigi Riz}
is currently a Research Scientist with the Technologies of Vision Unit at Fondazione Bruno Kessler, Trento, Italy. He received the B.Sc. and M.Sc. degrees in Computer Science from the University of Trento, Italy. His research interests include 3D scene understanding, 3D scene reconstruction, and computer vision.
\end{IEEEbiography}

\begin{IEEEbiography}[{\includegraphics[width=1in,height=1.25in,clip,keepaspectratio]{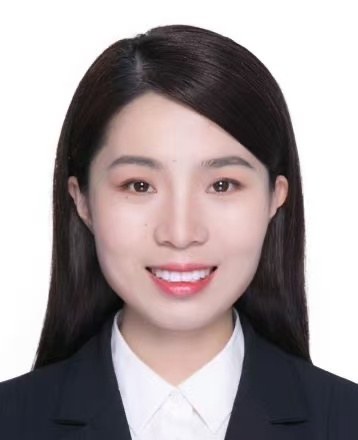}}]{Juan Liu}
is currently a Lecturer at Beijing Forestry University, Beijing, China. She received the Ph.D. degree in Systems Theory from the University of Chinese Academy of Sciences, Beijing, China, in 2019. Her research interests include general artificial intelligence, complex networks, and image recognition.
\end{IEEEbiography}

\begin{IEEEbiography}[{\includegraphics[width=1in,height=1.25in,clip,keepaspectratio]{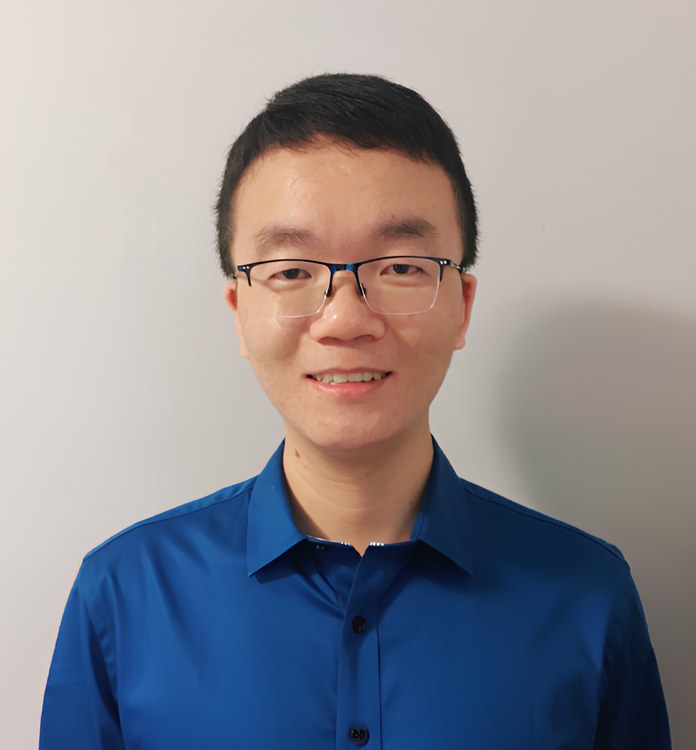}}]{Xiaoshui Huang}
is currently an Assistant Professor at Shanghai Jiao Tong University, Shanghai, China. His research interests include general artificial intelligence, 3D computer vision, foundation models, and AI applications in healthcare.
\end{IEEEbiography}

\begin{IEEEbiography}[{\includegraphics[width=1in,height=1.25in,clip,keepaspectratio]{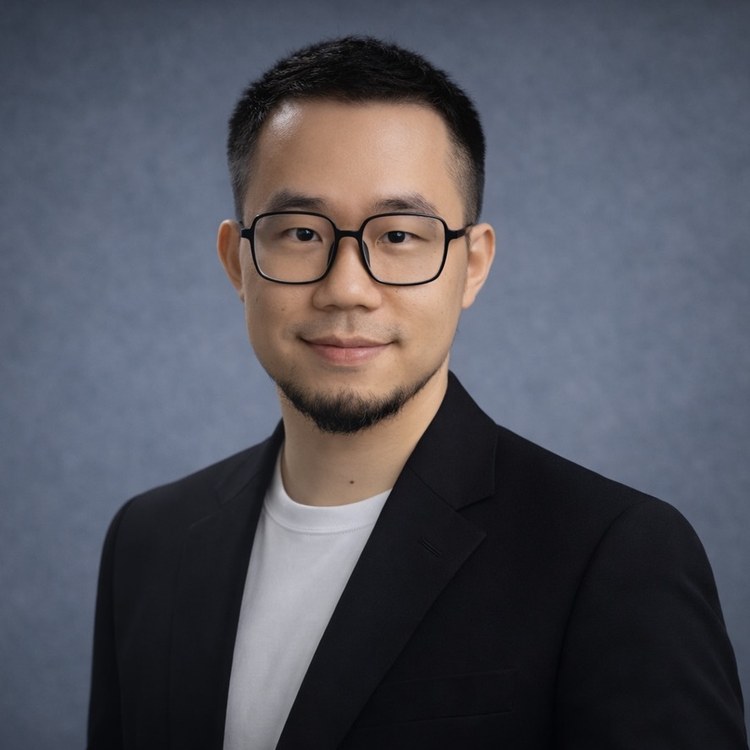}}]{Xu Zheng}
(Student Member, IEEE) is currently a Ph.D. candidate in the Artificial Intelligence Thrust at The Hong Kong University of Science and Technology (Guangzhou), Guangzhou, China. He received the B.E. and M.S. degrees from Northeastern University, China. His research interests include multimodal large models, multimodal learning, sensing, and perception.
\end{IEEEbiography}

\begin{IEEEbiography}[{\includegraphics[width=1in,height=1.25in,clip,keepaspectratio]{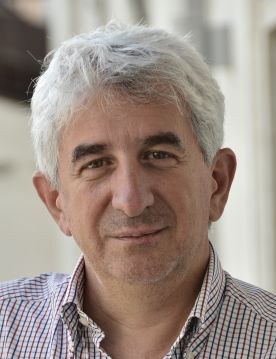}}]{Nicu Sebe}
(Senior Member, IEEE) is currently a Professor at the University of Trento, Trento, Italy, where he leads research in multimedia information retrieval and human behavior understanding in the Multimedia and Human Understanding Group (MHUG). He was the General Co-Chair of ACM Multimedia 2013 and the Program Chair of ACM Multimedia 2007 and 2011, ECCV 2016, ICCV 2017, and ICPR 2020. He is a Fellow of the International Association for Pattern Recognition.
\end{IEEEbiography}

\begin{IEEEbiography}[{\includegraphics[width=1in,height=1.25in,clip,keepaspectratio]{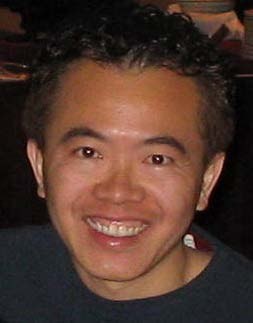}}]{Ming-Hsuan Yang}
(Fellow, IEEE) is currently a Professor of Electrical Engineering and Computer Science at the University of California, Merced, CA, USA. He served as a Program Co-Chair of the IEEE International Conference on Computer Vision (ICCV) in 2019 and as a General Co-Chair of ACCV 2016. He received the Longuet-Higgins Prize in 2023, the NSF CAREER Award in 2012, and the Google Faculty Award in 2009. He is a Fellow of the IEEE, ACM, AAAI, and AAAS.
\end{IEEEbiography}

\begin{IEEEbiography}[{\includegraphics[width=1in,height=1.25in,clip,keepaspectratio]{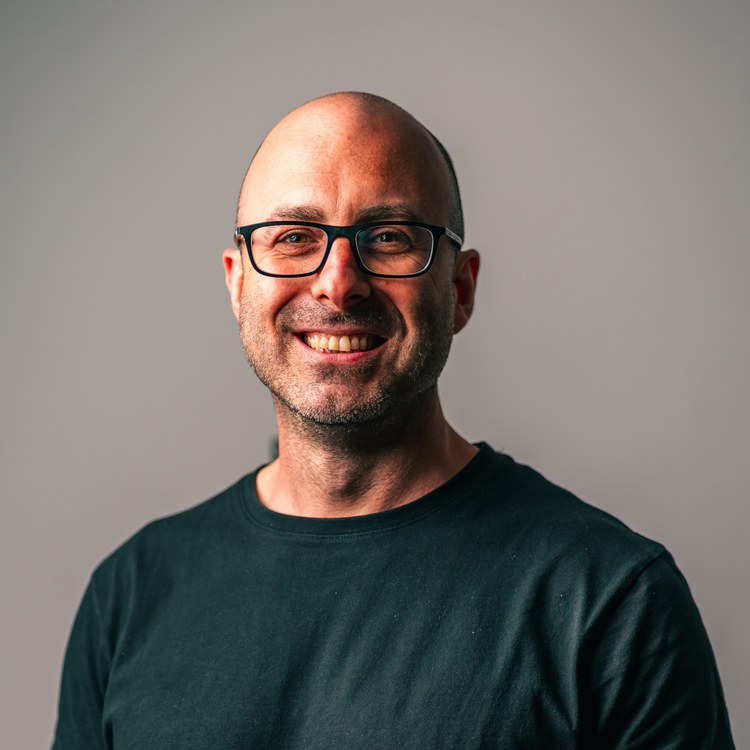}}]{Fabio Poiesi}
is currently the Head of the Technologies of Vision Unit at Fondazione Bruno Kessler, Trento, Italy. His research interests include 3D scene understanding, vision-language models, vision for robot manipulation, and robot learning. He is an Associate Editor of \textit{Pattern Recognition} and has served as an Area Chair for CVPR. He is an ELLIS Member.
\end{IEEEbiography}